\documentclass[]{interact}

\usepackage{epstopdf}
\usepackage{subfigure}

\usepackage[numbers,sort&compress]{natbib}
\bibpunct[, ]{[}{]}{,}{n}{,}{,}
\renewcommand\bibfont{\fontsize{10}{12}\selectfont}

\usepackage{fix-cm}
\usepackage{xcolor}
\usepackage{enumitem}
\usepackage{hyperref}
\usepackage{graphicx}
\usepackage[most]{tcolorbox}
\usepackage{float} 

\theoremstyle{plain}

\theoremstyle{definition}

\theoremstyle{remark}

\newtcolorbox{scenebox}[1][]{
  colback=gray!8,
  colframe=black,
  title=#1,
  fonttitle=\bfseries,
  boxrule=0.4pt,
  arc=2mm,
  left=2mm,
  right=2mm,
  top=1.5mm,
  bottom=1.0mm,
}

\usepackage{tikz}
\usetikzlibrary{shapes.geometric, arrows.meta, positioning}

\tikzset{
  step1/.style={
    rectangle, rounded corners,
    minimum width=3.8cm, minimum height=1.2cm,
    text centered, draw=black, fill=blue!10, font=\small
  },
  step2/.style={
    rectangle, rounded corners,
    minimum width=3.8cm, minimum height=1.2cm,
    text centered, draw=black, fill=green!10, font=\small
  },
  step3/.style={
    rectangle, rounded corners,
    minimum width=3.8cm, minimum height=1.2cm,
    text centered, draw=black, fill=orange!15, font=\small
  },
  myarrow/.style={
    thick,->,>=Stealth
  },
  phaselabel1/.style={
    rectangle, rounded corners,
    minimum width=4.4cm,
    text centered, draw=black, fill=blue!10,
    font=\bfseries\small, inner sep=8pt
  },
  phaselabel2/.style={
    rectangle, rounded corners,
    minimum width=4.4cm,
    text centered, draw=black, fill=green!10,
    font=\bfseries\small, inner sep=6pt
  },
  phaselabel3/.style={
    rectangle, rounded corners,
    minimum width=4.8cm,
    text centered, draw=black, fill=orange!15,
    font=\bfseries\small, inner sep=6pt
  }
}

\begin{document}


\title{Recomposed realities: animating still images via patch clustering and randomness
}

\author{
\name{Markus Juvonen\textsuperscript{a}\thanks{CONTACT Markus Juvonen. Email: markus.juvonen@helsinki.fi} and Samuli Siltanen\textsuperscript{a}}
\affil{\textsuperscript{a} University of Helsinki, Helsinki, Finland;} 
}

\maketitle

\begin{abstract}
We present a patch-based image reconstruction and animation method that uses existing image data to bring still images to life through motion. Image patches from curated datasets are grouped using k-means clustering and a new target image is reconstructed by matching and randomly sampling from these clusters. This approach emphasizes reinterpretation over replication, allowing the source and target domains to differ conceptually while sharing local structures. 
\end{abstract}


\section{Introduction}

From ancient mosaics and collage techniques used thousands of years ago, repourposing small pieces of material to create new works of art has a long history. In today's digital world, where millions of images are uploaded on various platforms daily, we can find a new way of recycling these visual data to create something new.

In this work, we introduce a process that reuses image data in order to bring any image to life in video format. The central idea is to reconstruct an image using patches drawn from a selected dataset, combining clustering techniques with an element of randomness. We choose the data to deliberately depart from the original context of the reconstructed image prioritizing reinterpretation over replication.

On the one hand, we will present the process in a mathematical framework. On the other hand, we will see how the choices we make can add new layers of meaning and depth to this process.

\subsection{Natural image subset}

Modern digital cameras produce color images with tens of megapixels. Even compressed images used online typically have at least one megapixel (one million pixels). Each pixel usually has 8-bit color depth per channel, giving 256 possible values per color channel. To simplify, let us only consider grayscale images with a single 8-bit channel. A one-megapixel grayscale image can have $2^{8^{1000000}}=2^{8000000}$ possible combinations of pixel values. This is an astronomically large number. 

\begin{figure}
    \centering
    \includegraphics[width=1\linewidth]{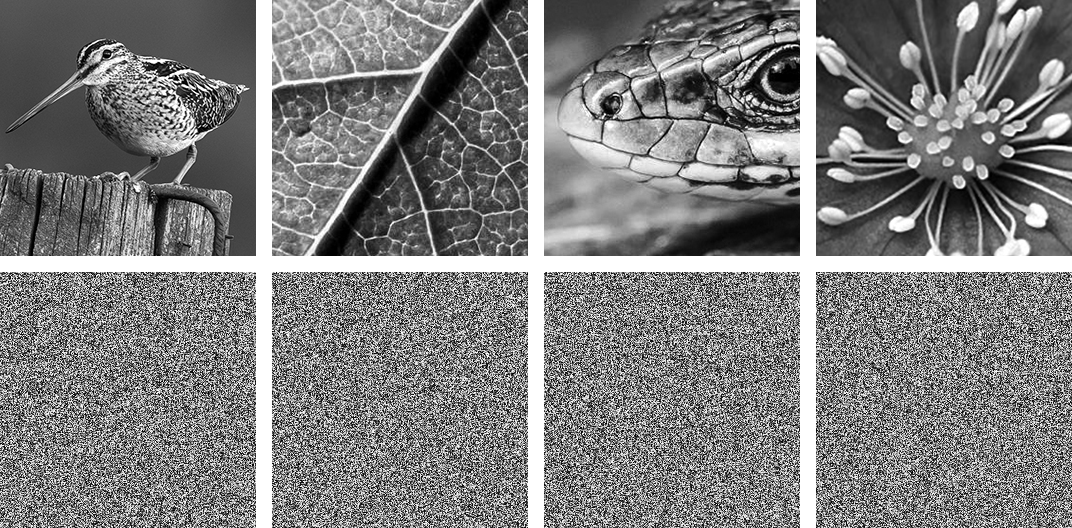}
    \caption{A selection of 4 natural images and 4 images with randomly chosen pixel values from a uniform distribution. Can you guess which are which?}
    \label{fig:nature_random}
\end{figure}

However, only a tiny fraction of these combinations would resemble natural photographs. Most randomly generated images of this size would appear as meaningless noise. We bet you can distinguish between the natural images and the random noise images in Figure \ref{fig:nature_random}. That’s because our perception relies on familiar structures such as edges, textures, and shapes that our brains are trained to recognize. Small changes in a few pixels are often imperceptible, reinforcing the idea that meaningful images occupy a small region within this vast space of possible images. The subspace of natural images has been researched in various ways in the last 40 years. From statistical approaches \cite{mumford2001stochastic,simoncelli2001natural} to geometric \cite{peyre2009manifold,lee2003nonlinear} and topological \cite{Carlsson20081} approaches, they all focus on locality and small parts of images rather than trying to understand the structure of large images at once, as the features and details found are often scalable in the end. Before taking a closer look at the patch-based image representation, let us take a step back and clarify what we mean by natural images. 

Caselles et al. \cite{caselles1997}, define natural images as any photograph taken of any scene with a regular camera. Others, like
Hyvärinen et al. \cite{hyvarinen2009natural} on the other hand, make the connection between natural images and the human visual system by defining natural images as some set that they believe has similar statistical structure to which our visual system is adapted to. This definition leads to a clear distinction between purely natural scenes and scenes that include the marks of human engineering, due to the fact that our visual system developed during a long period of time in which our environment was much different from today's modern world.

We will include man-made objects (e.g. buildings, streets, vehicles) in our definition of natural images because we are inclined to think of humans as part of the natural world. Organic and engineered scenes exhibit distinct structural characteristics, reflecting the contrast between patterns shaped by evolution and those imposed through human intention. Although we include man-made scenes in our natural image definition, we purposely separate these from nature images in our workflow and use this contrast as a guiding motif that shapes the way the final results are perceived and cognitively interpreted.

\subsection{Patch-based reconstruction approach}\label{patch_intro}

Using images as building blocks for digital mosaics is not a new idea. This idea started to arise in the first half of the 1990's as images started to be converted into digital form and computing power of the modern computer increased. However, the first photo mosaics used entire images (often downsampled) as the building blocks of a new image. A key difference behind this project is to use patches of images from a selected data set to reconstruct a given target image. One possibility could be to reconstruct a modern cityscape from a historical dataset of old city images. Or we could use the old city images to reconstruct a nature photograph as we did in the video for the Bridges 2023 Conference short film festival, which we will discuss in more detail in Section \ref{results}. One could also use nature photographs to construct an image picturing a urban setting with buildings, vehicles, etc. There is something compelling about uncovering shared patterns in images drawn from vastly different environments.

Patch-based methods, that decompose images into smaller overlapping or non-overlapping patches, typically square and of uniform size, have been very effective in several image processing tasks \cite{Alkinani2017,Karimi2016}, such as denoising \cite{elad2006}, deblurring \cite{dabov2008image}, inpainting \cite{barnes2009patchmatch}, texture synthesis \cite{efros2001quilting} and image segmentation \cite{coupe2011patch}. One significant advantage of patch-based approaches is their ability to effectively capture detailed local features while remaining computationally manageable. By analyzing small image patches, these methods exploit the redundancy in natural images.
 
The choice of patch size plays a key role in these approaches. Patch sizes commonly range from $5\times5$ to $32\times32$ depending a bit on the specific case of application, some of the methods adapting the patch size to different images or even within regions of images with different levels of detail. Large patches are computationally heavier but might capture larger structures better than small patches. You definitely want to select patches that are at least as large as the smallest structures you want to capture. 
 
We experimented with quite large patches (up to $128\times128$), as we welcome some clearly distinguishable structures from the original data in the final reconstructions even if it is computationally more costly, and we might not get the closest possible reconstruction to the original.

\begin{figure}[htbp]
\centering
\resizebox{1\textwidth}{!}{%
\begin{tikzpicture}[node distance=0.5cm and 1.3cm]

  \node (p1label) [phaselabel1] at (0, 0.8) {\shortstack{Phase 1\\[0.5ex] Patch Extraction\\ \& Clustering}};
  \node (s1) [step1, below=0.3cm of p1label] {1. Collect image dataset};
  \node (s2) [step1, below=of s1] {2. Extract image patches};
  \node (s3) [step1, below=of s2] {3. Center the patches};
  \node (s4) [step1, below=of s3] {4. Cluster similar patches};

  \node (p2label) [phaselabel2, right=1.5cm of p1label] {\shortstack{Phase 2\\[0.5ex] Patch Matching}};
  \node (s5) [step2, below=0.4cm of p2label] {5. Select target image};
  \node (s6) [step2, below=of s5] {6. Divide into patches};
  \node (s7) [step2, below=of s6] {7. Match to closest cluster};
  \node (s8) [step2, below=of s7] {8. Replace with random member};

  \node (p3label) [phaselabel3, right=1.5cm of p2label] {\shortstack{Phase 3\\[0.5ex] Animation}};
  \node (s9) [step3, below=0.4cm of p3label] {9. Select new randoms};
  \node (s10) [step3, below=of s9] {10. Repeat for more frames};
  \node (s11) [step3, below=of s10] {11. Assemble into video};

  \draw [myarrow] (s1) -- (s2);
  \draw [myarrow] (s2) -- (s3);
  \draw [myarrow] (s3) -- (s4);
  \draw [myarrow] (s5) -- (s6);
  \draw [myarrow] (s6) -- (s7);
  \draw [myarrow] (s7) -- (s8);
  \draw [myarrow] (s9) -- (s10);
  \draw [myarrow] (s10) -- (s11);

  \draw [myarrow] (s4.east) to[out=20, in=180] (s5.west);
  \draw [myarrow] (s8.east) to[out=40, in=180] (s9.west);

\end{tikzpicture}
}
\caption{Overview of the patch-based image reconstruction workflow. The process is divided into three phases: patch extraction \& clustering, patch matching, and animation.}
\label{fig:patch_pipeline}
\end{figure}
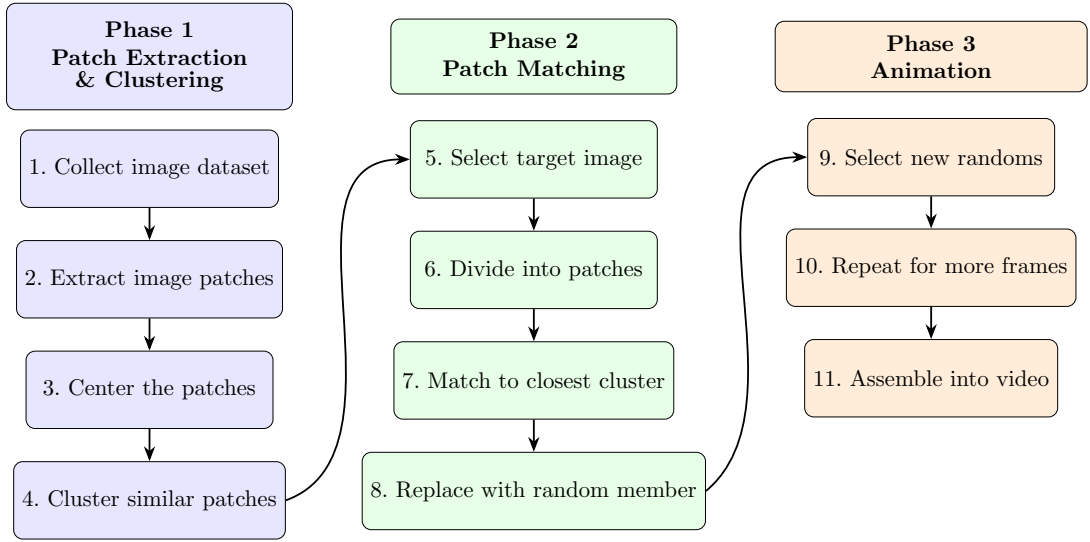

An overview of the workflow we use can be seen in Figure \ref{fig:patch_pipeline}. We have broken the workflow down into multiple steps that are divided into three phases. These steps will be explained in detail in Section 2. In Section 3 we show some results of this process and in Section 4 we discuss the outcomes and future possibilities.

\section{Materials and Methods}

For the sake of simplicity, we will restrict ourselves to 8-bit grayscale images of size $1024\times1024$ pixels in this work. This means that our image dimensions are neatly divisible by powers of 2 and will have 256 possible grayscale values in the range of 0 to 255, where 0 represents black and 255 represents white.

\subsection{Selection of the images (Phase 1 Step 1)}

We used two data sets. A selection of historical Helsinki images from the Helsinki City Museums open image database (\href{https://www.helsinkiphotos.fi/}{https://www.helsinkiphotos.fi/}) and some of the corresponding author's personal nature photographs posted on the social media platform Instagram. As we only use grayscale images of size 1024x1024 pixels, images were cropped to this size from the originals and converted to grayscale if not already in the selected format.

\begin{figure}
    \centering
    \includegraphics[width=1\linewidth]{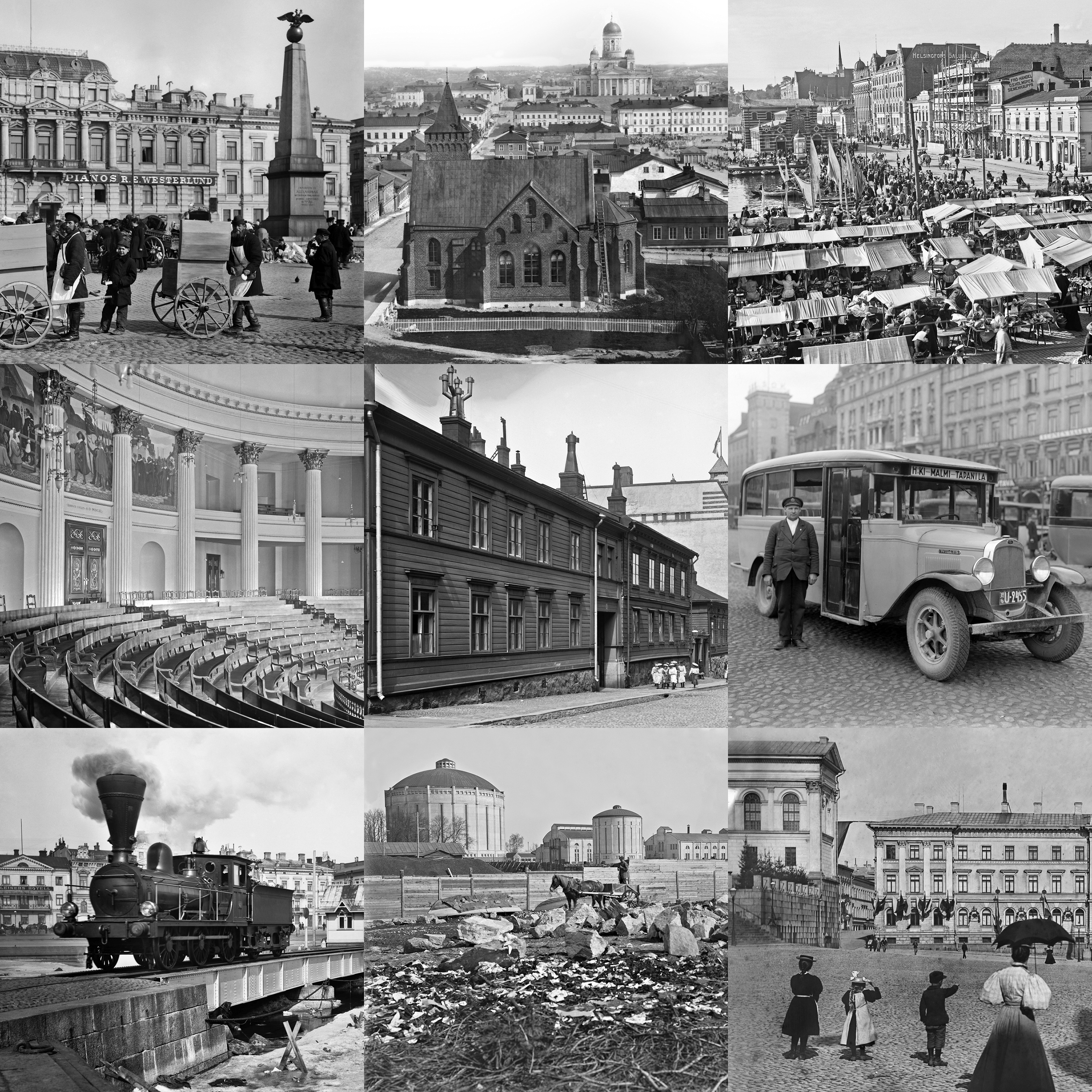}
    \caption{A selection of 9 out of 38 cropped images from the Old Helsinki data set used in this work.}
    \label{fig:3times3grid_stadiBW}
\end{figure}

\begin{figure}
    \centering
    \includegraphics[width=1\linewidth]{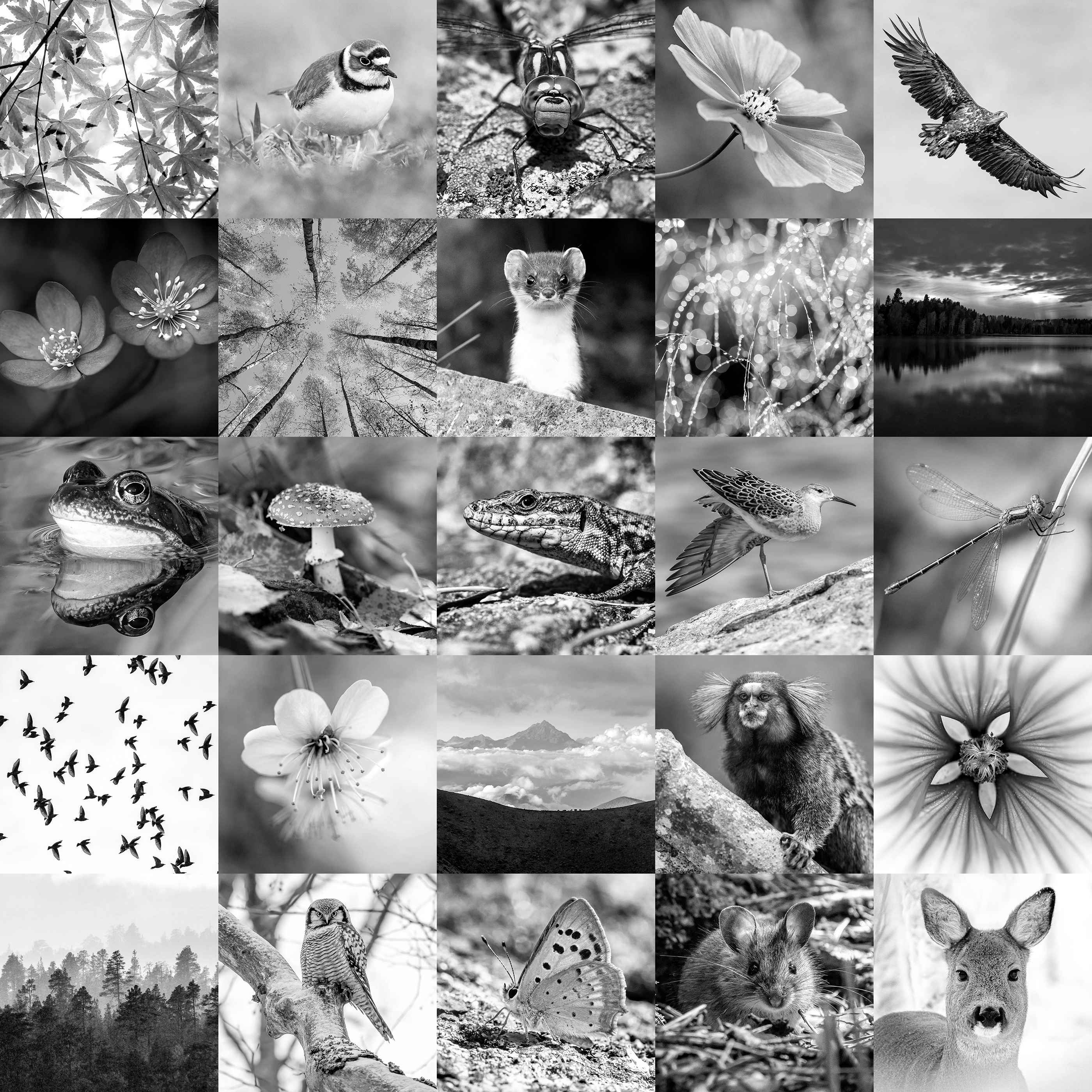}
    \caption{A selection of 25 out of 230 images from the Nature Image data set used.}
    \label{fig:5times5grid_instaBW}
\end{figure}

For the Old Helsinki data set we ended up with 38 final images cropped from the original images provided by the Helsinki City Museum. A selection of 9 of those images can be seen in Figure \ref{fig:3times3grid_stadiBW}. The second data set, the Nature Image data set, consists of 230 photographs. A selection of 25 images from this set can be seen in Figure \ref{fig:5times5grid_instaBW}.

\subsection{Extracting image patches (Phase 1 step 2)}

Consider a set of grayscale images:
\[
\mathcal{I} = \{I_{1}, I_{2}, \dots, I_{T}\},
\]
where each image \( I_{t} \in \mathbb{R}^{N \times N} \) and \( N = 2^a \) for some integer \( a \geq 2 \). We assume that each \( I_{t} \) is partitioned into square patches of size \( n \times n \), where \( n = 2^b \) for some integer \( b < a \). For each image, the number of patches is: 

\[
L = \left(\frac{N}{n}\right)^2.
\]

Combining patches from all \( T \) images, the total number of patches is:
\[
M = T \cdot L.
\]

We denote the entire set of extracted patches by:
\[
\mathcal{P} = \{\mathbf{p}_i \in \mathbb{R}^{n^2} : i = 1,2,\dots,M\},
\]
where each patch \(\mathbf{p}_i\) is reshaped into a vector of length \( n^2 \) and stored. Each image can also be partitioned into overlapping patches, which means that we can introduce a horizontal and vertical step size \( s \), which satisfies:
\[
s = 2^c, \quad \text{for some integer } c \leq b.
\] 

In this way, we can extract more patches from a smaller data set if necessary. We used overlapping patches for the Old Helsinki data set as we had fewer images than in the other data set. Especially with larger patch sizes, we can have more variance in the data by extracting overlapping patches.

The number of patches per image with overlapping patches is:
\[
L_{overlap} = \left(\frac{N - n}{s} + 1\right)^2.
\]

As we mentioned in Section \ref{patch_intro} the sizes of the patches are often chosen to be at least larger than the smallest structures visible in the images. Since we are working with grayscale images, we aim to distinguish at least between edges in different directions and varying brightness levels. As patch size increases, the variety of possible edge orientations and shapes within each patch also grows, which in turn requires a larger number of patches to adequately represent all these variations. This is often referred to as the curse of dimensionality.

If we use very small patches, we lose the ability to see the different textures and origins of the patches. For a more artistic approach we can increase the patch size more, as we are not only interested in getting the best matching reconstruction, but also to have thought-provoking results. We used patches ranging from $8\times8$ pixels up to $128\times128$ pixels in the results we present in Section \ref{results}.

\subsection{Centering the patches and K-means clustering (Phase 1 Steps 3 and 4)}

We use the k-means algorithm to divide the set of all patches $\mathcal{P}$ into \( k \) subsets that are called clusters. The algorithm minimizes the within-cluster sum of squares i.e. the variance.
For the clustering done in the video project we used the original k-means algorithm also called Lloyd's algorithm \cite{lloyd1982least}. Before running the clustering algorithm, the patches are typically centered.

\subsubsection*{Patch Centering (Phase 1 Step 3):}

Given the set of patches extracted from multiple images:
\[
\mathcal{P} = \{\mathbf{p}_i \in \mathbb{R}^{n^2} : i = 1,2,\dots,M\},
\]
each patch is individually centered by subtracting its mean pixel intensity. Thus, the centered patches are:
\[
\tilde{\mathbf{p}}_i = \mathbf{p}_i - \mu_i, \quad\text{with}\quad 
\mu_i = \frac{1}{n^2}\sum_{j=1}^{n^2}(\mathbf{p}_i)_j.
\]
This results in the centered patch set:
\[
\tilde{\mathcal{P}} = \{\tilde{\mathbf{p}}_1,\tilde{\mathbf{p}}_2,\dots,\tilde{\mathbf{p}}_M\}.
\]

\subsubsection*{K-means clustering (Phase 1 Step 4):}

The following clustering step partitions the centered patches \(\tilde{\mathcal{P}}\) into \( k \) distinct clusters $\mathcal{C} = \{C_1,C_2,\dots,C_k\}$, such that:

\[
\bigcup_{j=1}^{k} C_j = \tilde{\mathcal{P}}, \quad\text{and}\quad C_j\cap C_{j'} = \emptyset\text{ for } j\neq j'.
\]

This partitioning is obtained by minimizing the within-cluster sum of squared Euclidean distances:

\[
J(\mathcal{C},\mathbf{c}_1,\dots,\mathbf{c}_k) 
= \sum_{j=1}^{k}\sum_{\tilde{\mathbf{p}}\in C_j}\|\tilde{\mathbf{p}} - \mathbf{c}_j\|_2^2,
\]

where \(\mathbf{c}_j \in \mathbb{R}^{n^2}\) is the centroid (mean vector) of cluster \(C_j\), computed as:

\[
\mathbf{c}_j = \frac{1}{|C_j|}\sum_{\tilde{\mathbf{p}}\in C_j}\tilde{\mathbf{p}}.
\]

The clustering problem is solved iteratively through the following steps:

\begin{enumerate}
    \item \textbf{Initialization:} Randomly select initial cluster centroids \(\{\mathbf{c}_j\}_{j=1}^{k}\) from the set of patches.
    
    \item \textbf{Assignment Step:} Assign each patch \(\tilde{\mathbf{p}}_i\) to the nearest cluster centroid:
    \[
    C_j = \{\tilde{\mathbf{p}}_i : \|\tilde{\mathbf{p}}_i - \mathbf{c}_j\|_2 \leq \|\tilde{\mathbf{p}}_i - \mathbf{c}_{j'}\|_2,\;\forall j'\neq j\}.
    \]

    \item \textbf{Update Step:} Update each cluster centroid based on the current assignment:
    \[
    \mathbf{c}_j^{(\text{new})} = \frac{1}{|C_j|}\sum_{\tilde{\mathbf{p}}\in C_j}\tilde{\mathbf{p}}.
    \]

    \item \textbf{Convergence Check:} Repeat steps 2 and 3 until the cluster centroids stabilize or the change in the objective function \(J(\mathcal{C},\mathbf{c}_1,\dots,\mathbf{c}_k)\) falls below a pre-specified threshold \(\epsilon\):
    \[
    \|\mathbf{c}_j^{(\text{new})}-\mathbf{c}_j^{(\text{old})}\|_2<\epsilon,\quad j=1,\dots,k.
    \]
\end{enumerate}

After convergence, the final clusters \(C_j\) and their respective centroids \(\mathbf{c}_j\) constitute the clustering solution used in the reconstruction step. The k-means algorithm can only be guaranteed to find a local optimum \cite{Selim198481} and is thus often run multiple times with different starting conditions.

How to choose the number of clusters $k$? There are various methods to try to find the optimal $k$ \cite{j2020016,Pham2005}, but they may be ambiguous and often require that you run the clustering for many different values, and this can be very time consuming. Specific methods might also only work for some specific type of data well. The corresponding author has experimented with multiple $k$-values for clustering image patch data in his master's thesis \cite{juvonen2017patch} and in this work the most important thing is to have enough clusters to capture many different possible image features. We are working with large numbers of image patches, so we decided to have a large number of clusters rather than to underfit the data. If $k$ is too small, patches will be added to clusters where the mean might not really represent them well. If $k$ is large, we get more varied clusters, but if $k$ is too large, we might start to overfit and force similar patches into separate clusters and end up with many clusters with only one or a few representatives. If one of these clusters is selected to represent the new image patch in Phase 2, there is not much choice within the cluster, and that patch can become static in the video. The number of clusters $k$ was naturally chosen to be larger for small patches as the number of patches increases. For the Old Helsinki data set, the number of clusters ranged from $k=166$ for the $128\times128$ patches to $k=512$ for the $8\times8$ patches.

\begin{figure}[H]
    \centering
    \includegraphics[width=1\linewidth]{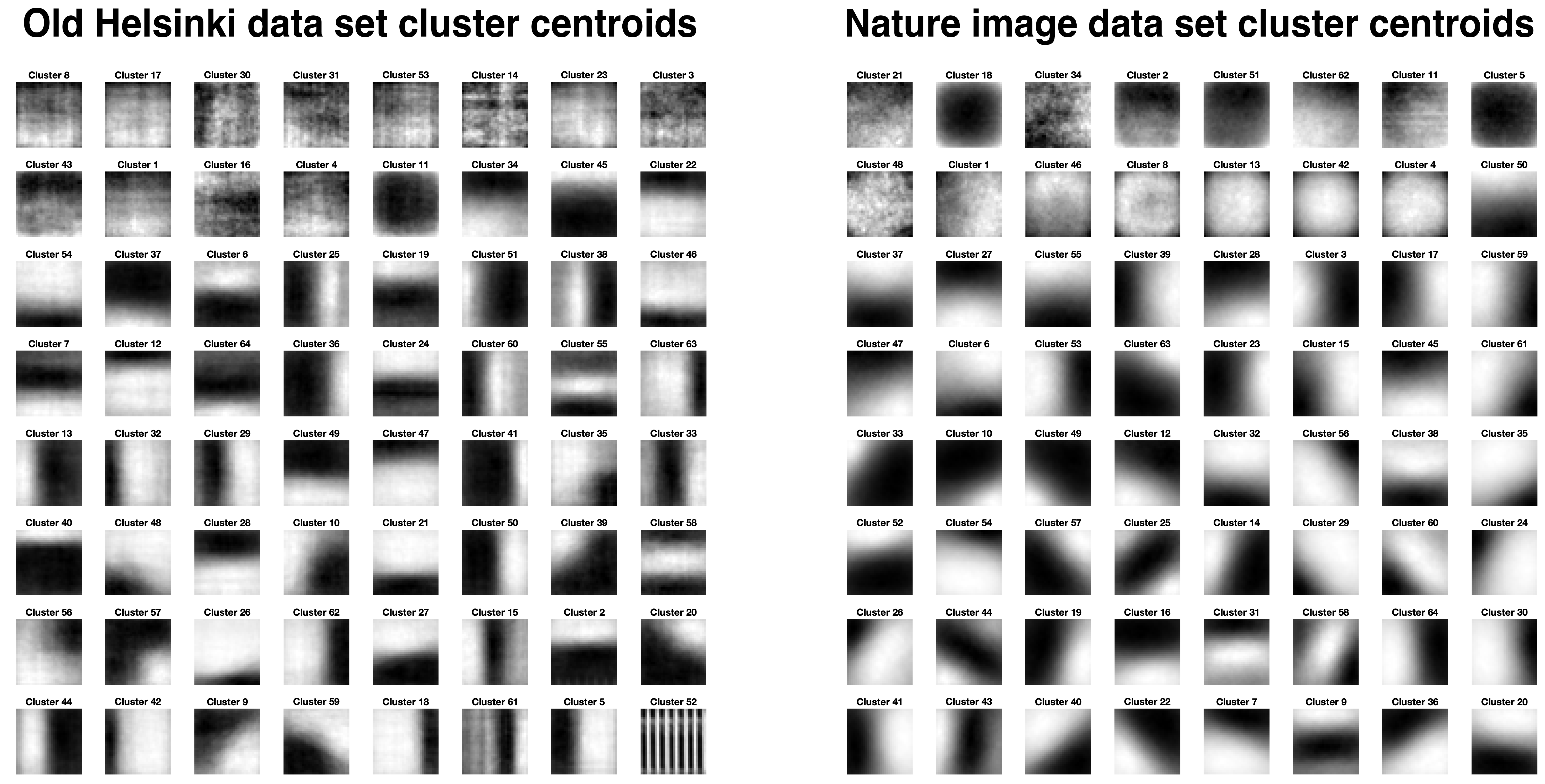}
    \caption{Cluster centroids from both data sets ordered from largest to smallest according to the amount of patches in the cluster (going row by row starting from the top-left). The patch sizes are $32\times32$ pixels and the amount of clusters is here 64 for both data sets. We see that the city data has much more horizontal and vertical edges than the nature set and vice versa the nature set seems to have more diagonal edges. The city set also has one cluster (bottom right in the left image) with higher frequency information.}
    \label{fig:Cluster_means_k64_instaBW}
\end{figure}

Figure \ref{fig:Cluster_means_k64_instaBW} shows the cluster centroids after k-means clustering of both data sets $32\times32$ sized patches without overlap into 64 clusters. This means that we had 38912 patches in the Old Helsinki data and 235520 patches in the nature image data. The city images clearly have more horizontal and vertical edges than the nature images. The nature image patches have more diagonal edges instead. We also see that the centroids after clustering the $32\times32$ patches from both data sets into 64 clusters do not really have many high-frequency features (except the last cluster of the Old Helsinki data). We can compare these centroids with the main components of both image patch sets using principal component analysis (PCA) seen in Figure \ref{fig:PCA_DCT_32size} in the appendix, with additional comparison to the 2D discrete cosine transform (DCT) basis functions and also the centroids for the $8\times8$ patches for $k=64$. The main reason for the missing high-frequency details in the k-means clustering is that k-means focuses predominantly on the largest differences between patches, which tend to be low-frequency (mean intensity or simple edges). High-frequency details are subtle, typically low-energy variations that easily get averaged out. This is one of the reasons why we don't use the actual centroids for reconstruction but real representative patches from the clusters.

\begin{figure}[H]
    \centering
    \includegraphics[width=1\linewidth]{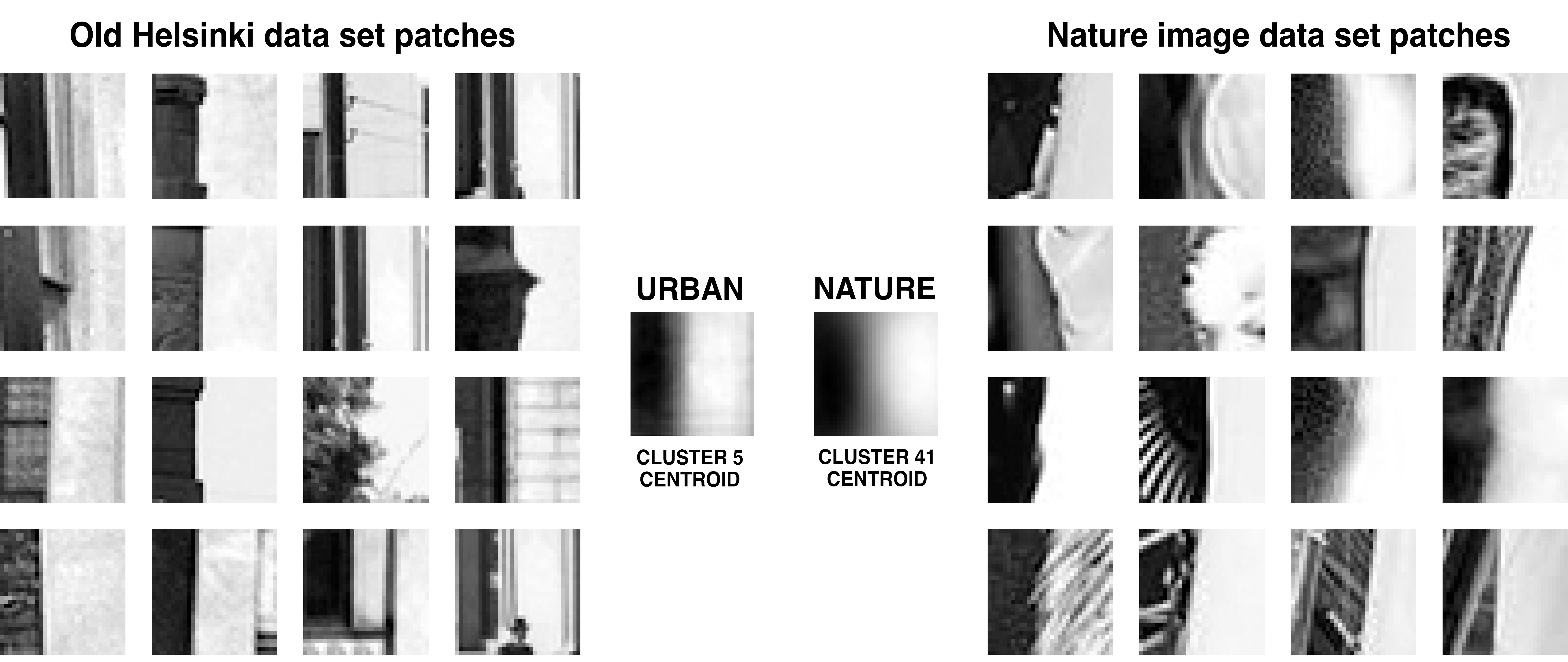}
    \caption{Example of 16 representatives selected from two clusters that have a similar looking centroid for the two different data sets. On the left we see the cityscape patches and on the right the nature image patches. Some of the patches look quite similar but some of the details are quite different.}
    \label{fig:Cluster_reps_k32_comparison}
\end{figure}

In Figure \ref{fig:Cluster_reps_k32_comparison} we take a closer look at a selection of 16 patches from two clusters that have a similar centroid with a prominent vertical edge for both data sets. In the city data we see, as expected, mostly parts of the sides of buildings with a vertical edge, but also a partial tree, a utility pole, and a hat-wearing gentleman walking in front of a building with a vertical edge. The nature patches reveal some fur and feather details and less strictly vertical edges with some variance in the curvature of the lines. This and the larger number of patches in this cluster (824 patches compared to the 106 patches in the urban cluster) explain why the centroid of the nature cluster averages out to a blurrier edge. 

\subsection{Reconstruction and random selection (Phase 2)}

Given a new grayscale image \( I' \in \mathbb{R}^{N \times N} \) where \( N = 2^a \), we partition it into patches as we did for the data set images before:
\[
\mathcal{P}' = \{\mathbf{q}_i \in \mathbb{R}^{n^2} : i=1,2,\dots,L\}.
\]

Each patch is centered:
\[
\tilde{\mathbf{q}}_i = \mathbf{q}_i - \mu'_i, \quad \text{where} \quad 
\mu'_i = \frac{1}{n^2}\sum_{j=1}^{n^2}(\mathbf{q}_i)_j.
\]

For each centered patch \(\tilde{\mathbf{q}}_i\), we select the closest cluster centroid
\(\mathbf{c}_j\) (obtained during the clustering of the original dataset patches):
\[
j^* = \arg\min_{1\leq j\leq k}\|\tilde{\mathbf{q}}_i - \mathbf{c}_j\|_2.
\]

Once the best matching cluster \(C_{j^*}\) is determined, we do not directly reconstruct the image using the centroid. Instead, we randomly select a representative $\mathbf{p}_{\text{rand}}$ from the original (non-centered) patches $C_{j^*}^\text{original}$ in the cluster. Since \(\mathbf{p}_{\text{rand}}\) is already an original patch with its natural intensity distribution, we reconstruct the patch \(\mathbf{q}_i\) directly as:
\[
\hat{\mathbf{q}}_i = \mathbf{p}_{\text{rand}}.
\]

This step avoids recentering or manually adding back any mean values, simplifying reconstruction and preserving natural intensity relationships from the dataset.

Finally, these reconstructed patches are assembled to form the reconstructed image \(\hat{I}' \in \mathbb{R}^{N\times N}\).\\

\subsection{Animation (Phase 3)}

After reconstructing the image $I'$ once and ending up with the first reconstruction $\hat{I}'$ we can simply repeat the random selection step several times to end up with slightly different variations and save each reconstruction as a separate image. We use random selection to make things more interesting and in our case to bring life to a still object. Perhaps we can view our newly built video frames as siblings that descend from the same process. They share the same blueprint and construction materials, but don't look exactly the same. Differences make things in general more interesting and exciting, and as human perception is sensitive to small changes, these changes evoke a sense of something evolving or moving. Similarly to the inherently random shimmering of light on water or a flame subtly flickering, we find ourselves captivated and mesmerized by it. 

 The final step in the described workflow is to turn the sequence of images or frames into a video clip. The number of frames and the frame rate of the video can be varied as desired.

\subsection{Histogram matching}
For a visually more consistent result to the original image $I'$, histogram matching \cite{alma9921839343506253, GONZALES1977111} can additionally be applied to each patch. The histogram matching operation transforms the histogram of an image to match the histogram of a reference image. In our case, we want to match the histograms of each $\hat{\mathbf{q}}_i$ to the corresponding reference patch $\mathbf{q}_i$.

\[
\hat{\mathbf{q}}_i^{\text{matched}} = \text{HistogramMatch}(\hat{\mathbf{q}}_i, \mathbf{q}_i).
\]

\begin{figure}
    \centering
    \includegraphics[width=1\linewidth]{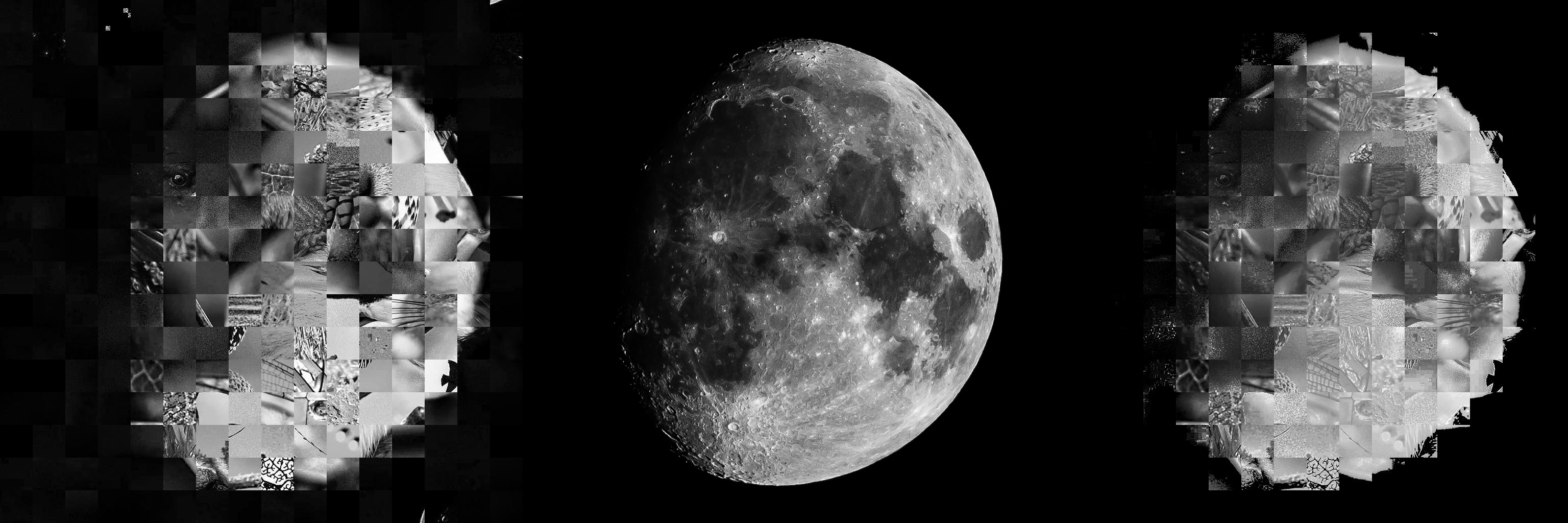}
    \caption{Moon image example with 64x64 sized patches. Left: Without histogram matching. Middle: original image. Right: with histogram matching for each patch.}
    \label{fig:moon_64_comparison}
\end{figure}

Figure \ref{fig:moon_64_comparison} shows the result of the use of histogram matching compared to the result without matching.

\section{Results} \label{results}

Presenting dynamic visual results inherently poses a challenge in written form, as essential temporal nuances captured in video are reduced to a few isolated snapshots. The static images can only provide limited insight into the continuous transformations crucial for a comprehensive understanding. In addition, the music accompaniment for a video makes a big difference. For a complete experience of our results, we warmly encourage the reader to view the accompanying video, where visuals and music blend seamlessly to reveal layers of meaning otherwise inaccessible. All computational steps in this work were performed using the MATLAB programming language and environment. The video editing steps used to create a video from still frames were performed using Adobe Premiere Pro.

\subsection{Video for the 2023 Bridges Short Film Festival}

Here we describe the key scenes of the video work named "Patch As Patch Can" that was presented at the 2023 Bridges Conference Short Film Festival in Halifax, Nova Scotia, Canada. Each scene is annotated with a timestamp range, illustrative still frames, and a brief interpretive commentary. The video was made for a general audience and focused on visual storytelling without a detailed explanation of each step of the process. It can be viewed on YouTube via the following link: \href{https://youtu.be/iu2HVdAQlh8?si=WyLQ1FN7XNX07eQy}{Patch As Patch Can}. The intrinsic video resolution is $1920\times1080$.

\begin{scenebox}{[0:07–0:19] Scene 1: Sneak peak of the result}
\begin{figure}[H]
  \centering
  \includegraphics[width=0.9\linewidth]{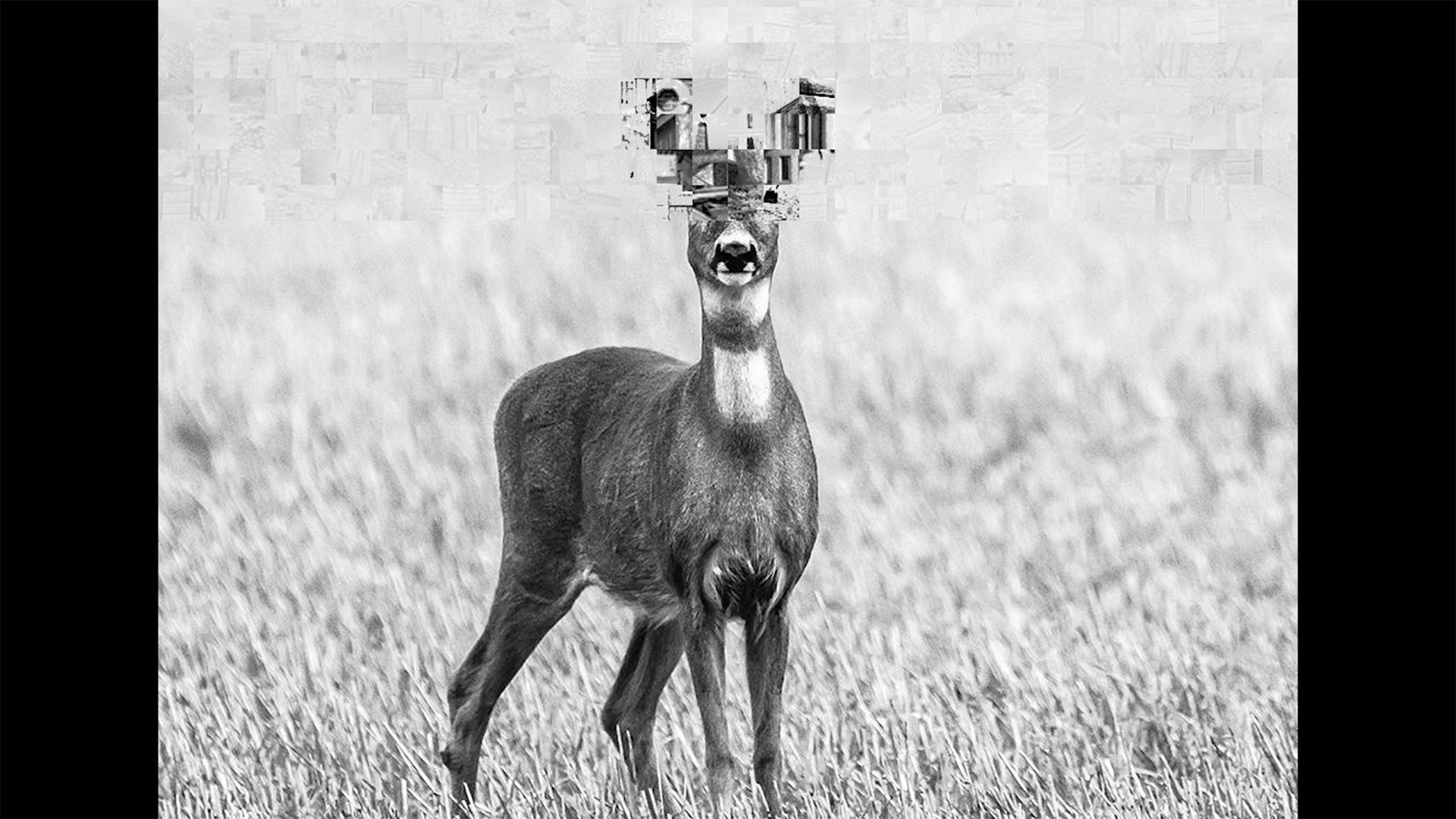}
  \caption{Still from Scene 1. The roe deer is replaced by an alternative, fragmented representation.}
\end{figure}

The opening scene begins with the image of a roe deer standing in a field. Slowly, from top to bottom, the image is replaced row by row by patches that somehow match the target but also change from frame to frame. This is a quick preview of the effect of this method.
\end{scenebox}

\vspace{0.5cm}

\begin{scenebox}{[0:21–0:42] Scene 2: Selecting an image database}
\begin{figure}[H]
  \centering
  \includegraphics[width=0.85\linewidth]{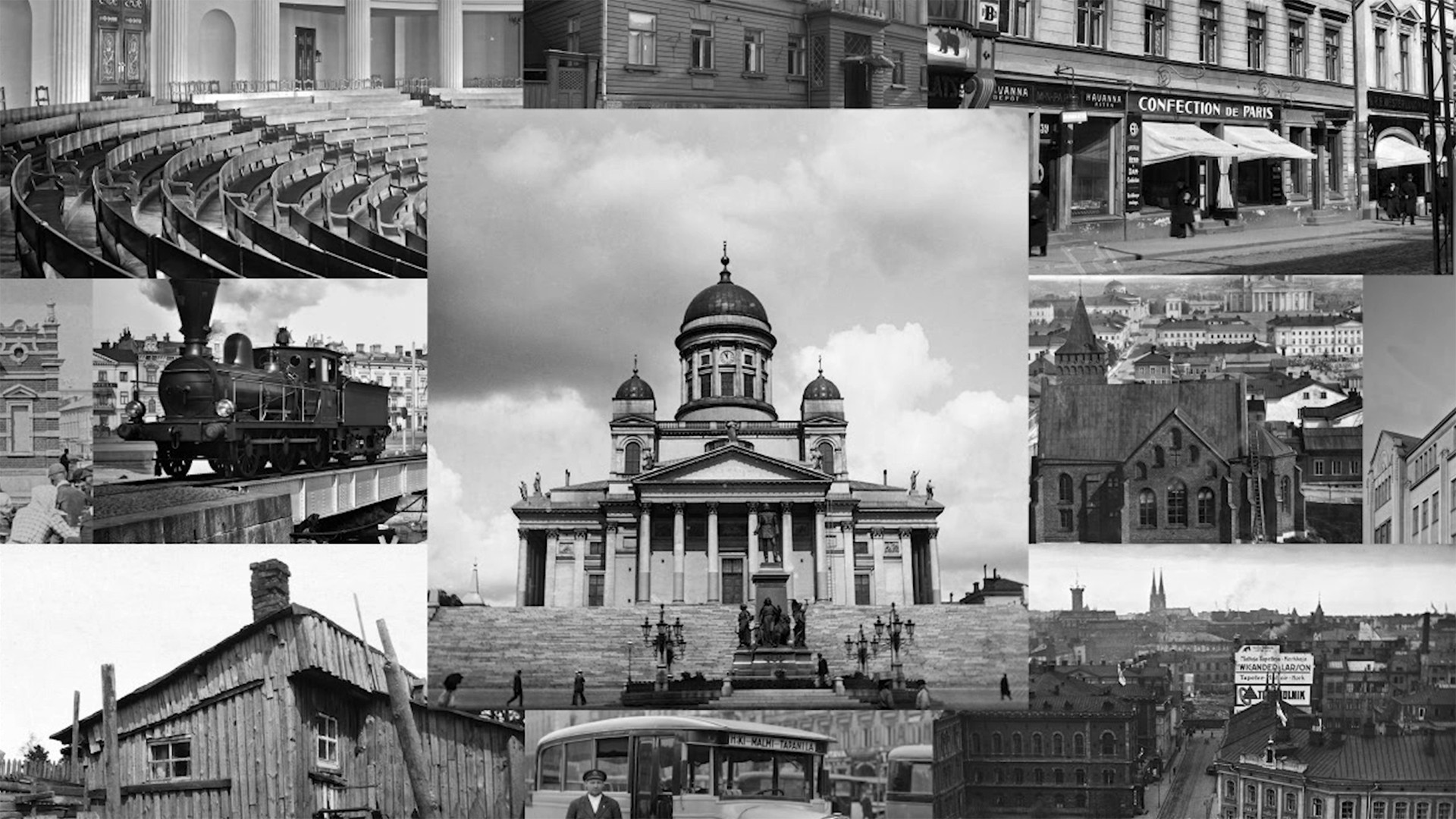}
  \caption{Still from Scene 2. A selection of images from the Old Helsinki data set.}
\end{figure}

The viewer is presented with the visually appealing Old Helsinki image set, courtesy of the Helsinki City Museum. The pace of the images appearing in the frame is adjusted to flow with the music.
\end{scenebox}

\begin{scenebox}{[0:43–1:08] Scene 3: Breaking the images into patches}
\begin{figure}[H]
  \centering
  \includegraphics[width=0.85\linewidth]{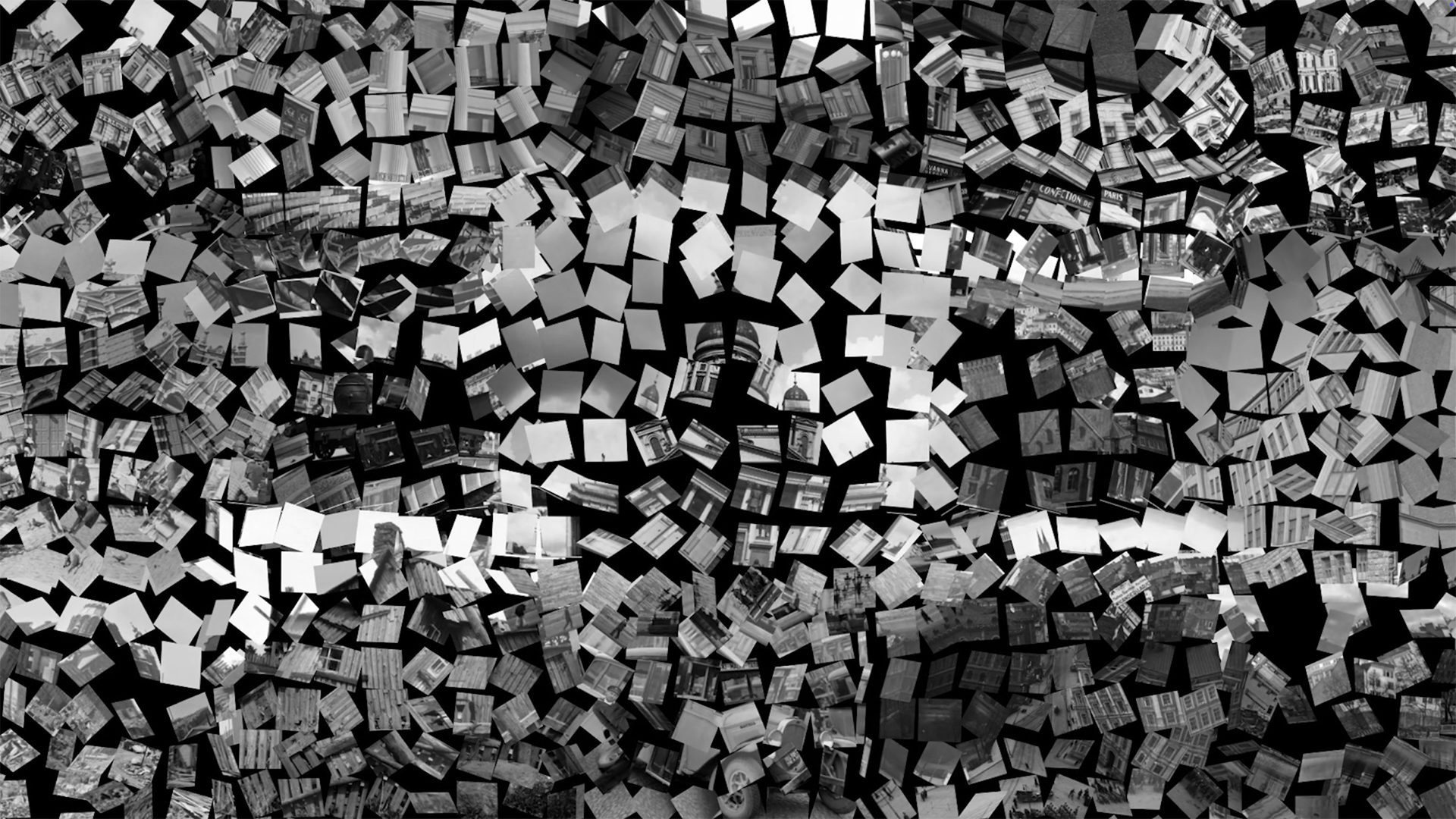}
  \caption{Still from Scene 3. The image set explodes into square patches.}
\end{figure}

The images "explode". Music brings tension that is released by a high piano note at the exact moment the images explode into pieces and scatter in every direction out of frame.
\end{scenebox}

\begin{scenebox}{[1:09–1:41] Scene 4: Sorting the patches into groups}
\begin{figure}[H]
  \centering
  \includegraphics[width=0.9\linewidth]{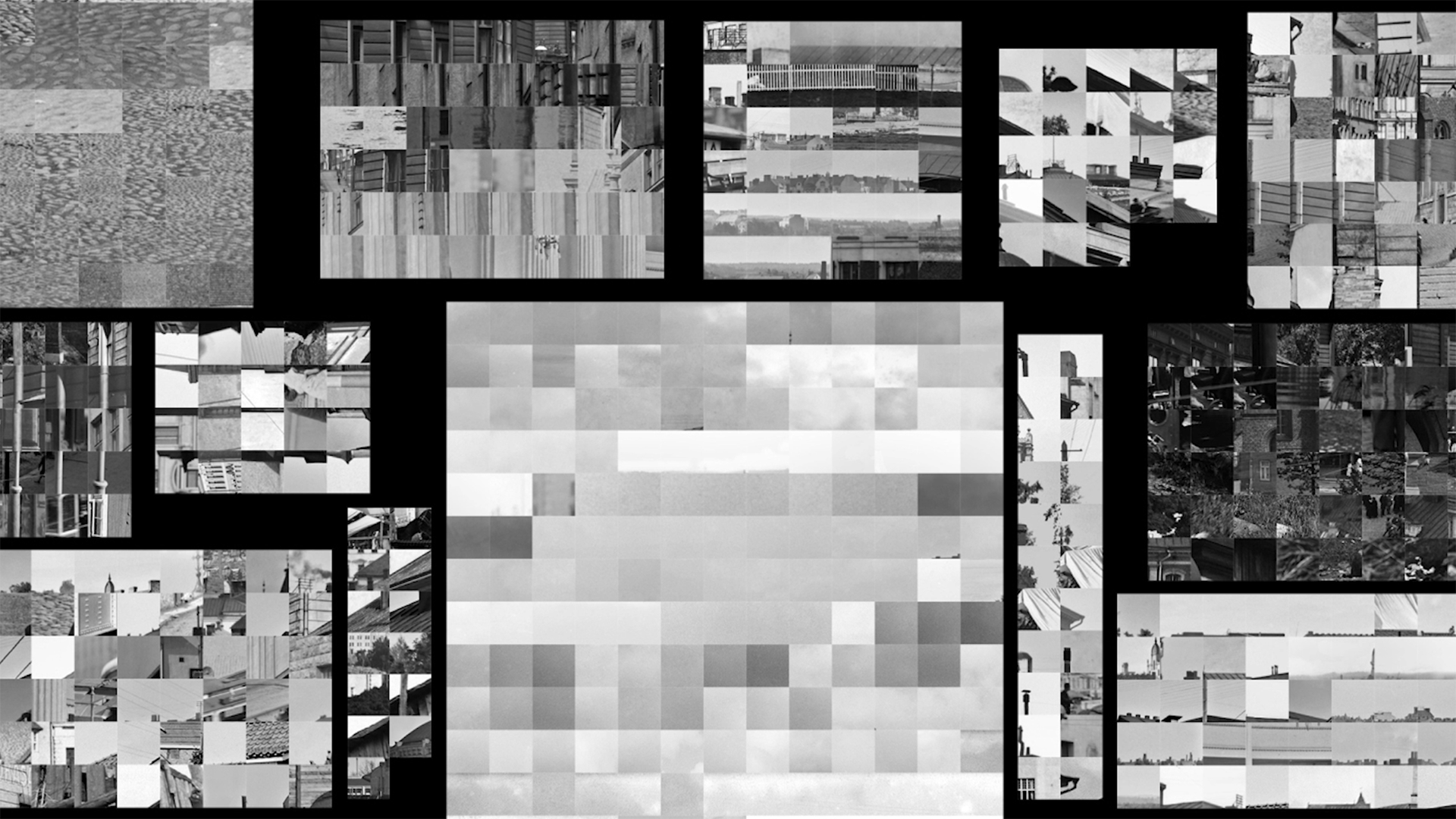}
  \caption{Still from Scene 4. Groups of actual image patches sorted into groups by k-means clustering.}
\end{figure}

Reversing the explosion effect we start from the patches reappearing in the frame from all sides but we see that they start to arrange themselves neatly into groups with similar patches. These are some of the actual representatives from some actual clusters selected to be shown as examples here. 

\end{scenebox}

\begin{scenebox}{[1:42–1:53] Scene 5: Selecting a reference image and partitioning it}
\begin{figure}[H]
  \centering
  \includegraphics[width=0.9\linewidth]{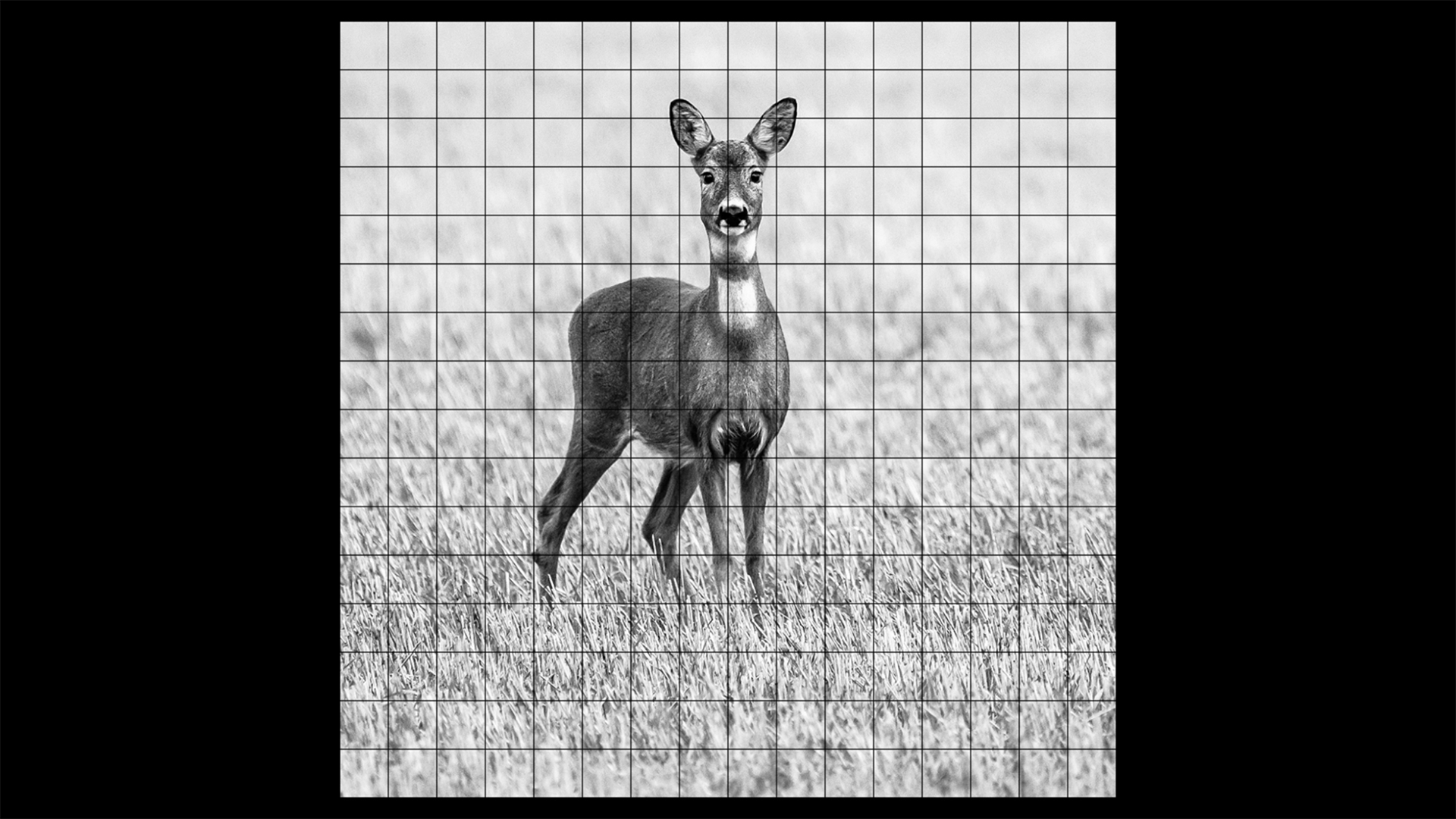}
  \caption{Still from Scene 5. The reference image picturing a roe deer is split into patches.}
\end{figure}

We see the familiar roe deer from the opening scene again. We see it getting partitioned into equally sized square patches.

\end{scenebox}

\begin{scenebox}{[1:54–2:15] Scene 6: Replace the reference patches with best matching group}
\begin{figure}[H]
  \centering
  \includegraphics[width=0.9\linewidth]{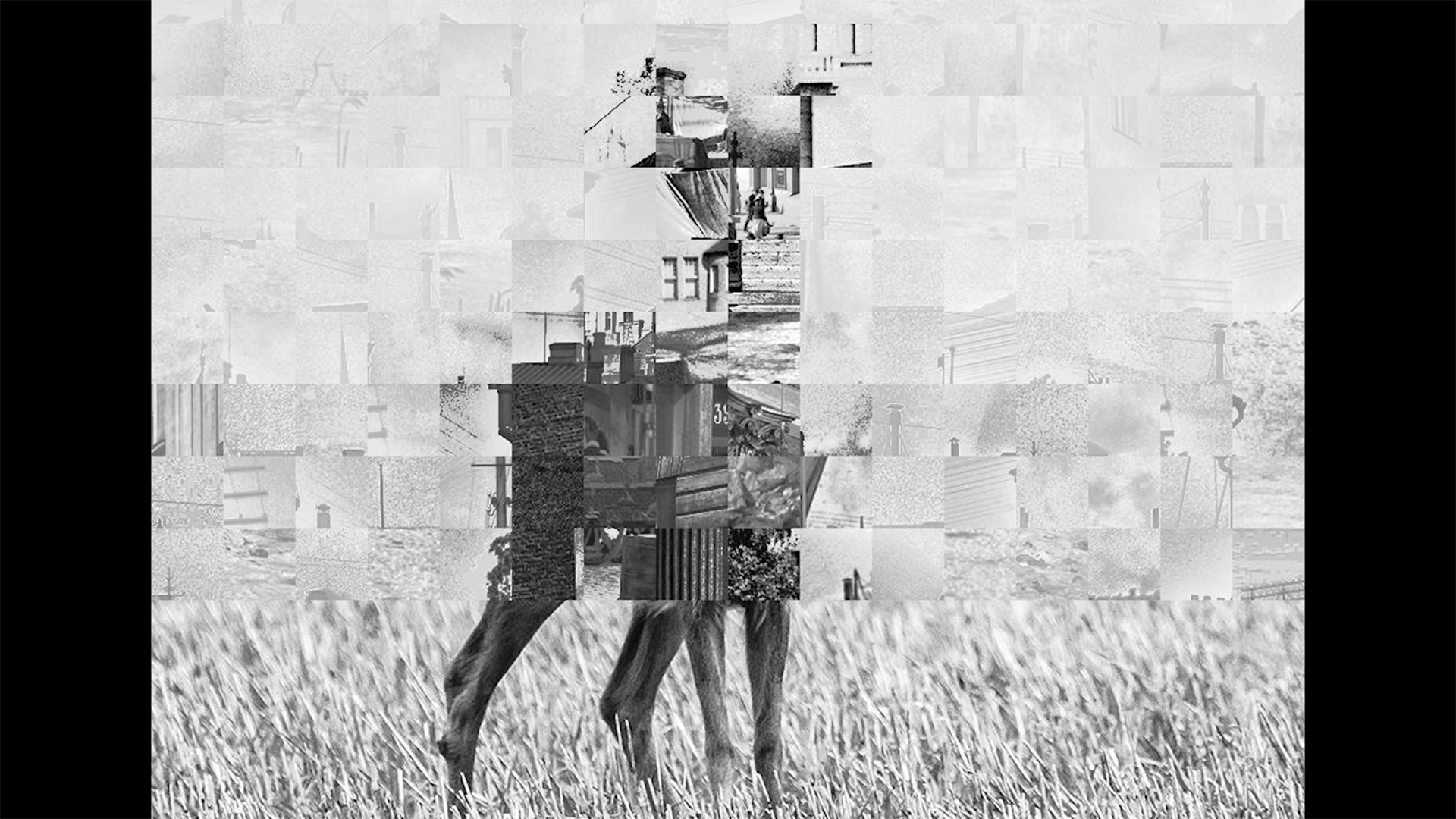}
  \caption{Still from Scene 6. The reference image is replaced row by row by 64x64 sized patches.}
\end{figure}

In this scene we see the end result again but now with more information about what is happening.

\end{scenebox}

\begin{scenebox}{[2:16–2:56] Scene 7: Differences in patch size}
\begin{figure}[H]
  \centering
  \includegraphics[width=0.9\linewidth]{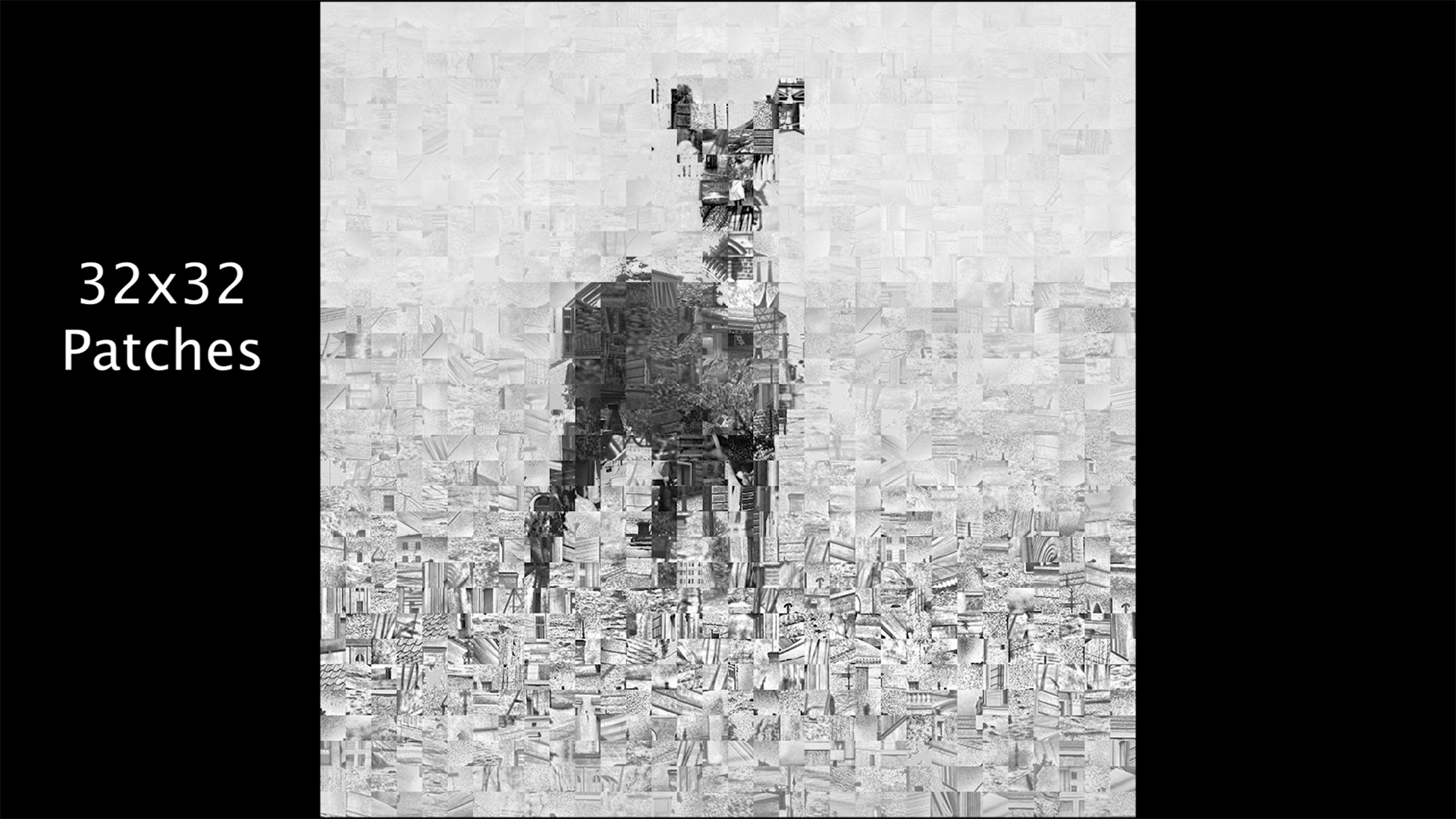}
  \caption{Still from Scene 7. Reconstruction with 32x32 patches from the Old Helsinki data set.}
\end{figure}

This scene shows the difference the patch size makes in the resulting reconstructions. Starting with $64\times64$ patches, we zoom out and switch to $32\times32$ patches and then to $16\times16$ patches. In the end, we zoom in on the deer and switch to $8\times8$ patches.

\end{scenebox}

\begin{scenebox}{[2:57–3:58] Scene 8: Different data set and blending modes}

\begin{figure}[H]
  \centering
  \begin{minipage}[b]{0.45\linewidth}
    \includegraphics[width=\linewidth]{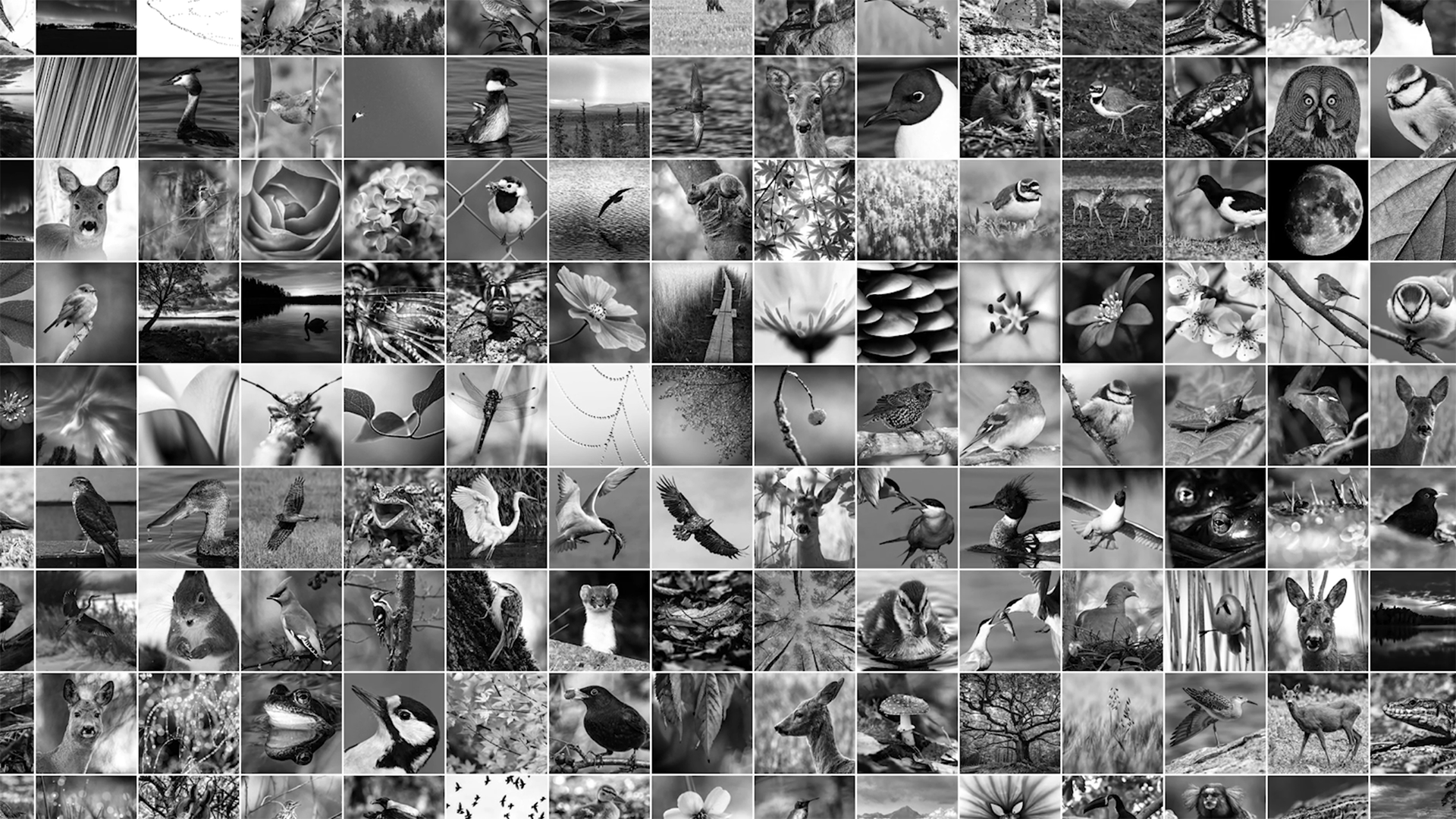}
  \end{minipage}
  \hfill
  \begin{minipage}[b]{0.45\linewidth}
    \includegraphics[width=\linewidth]{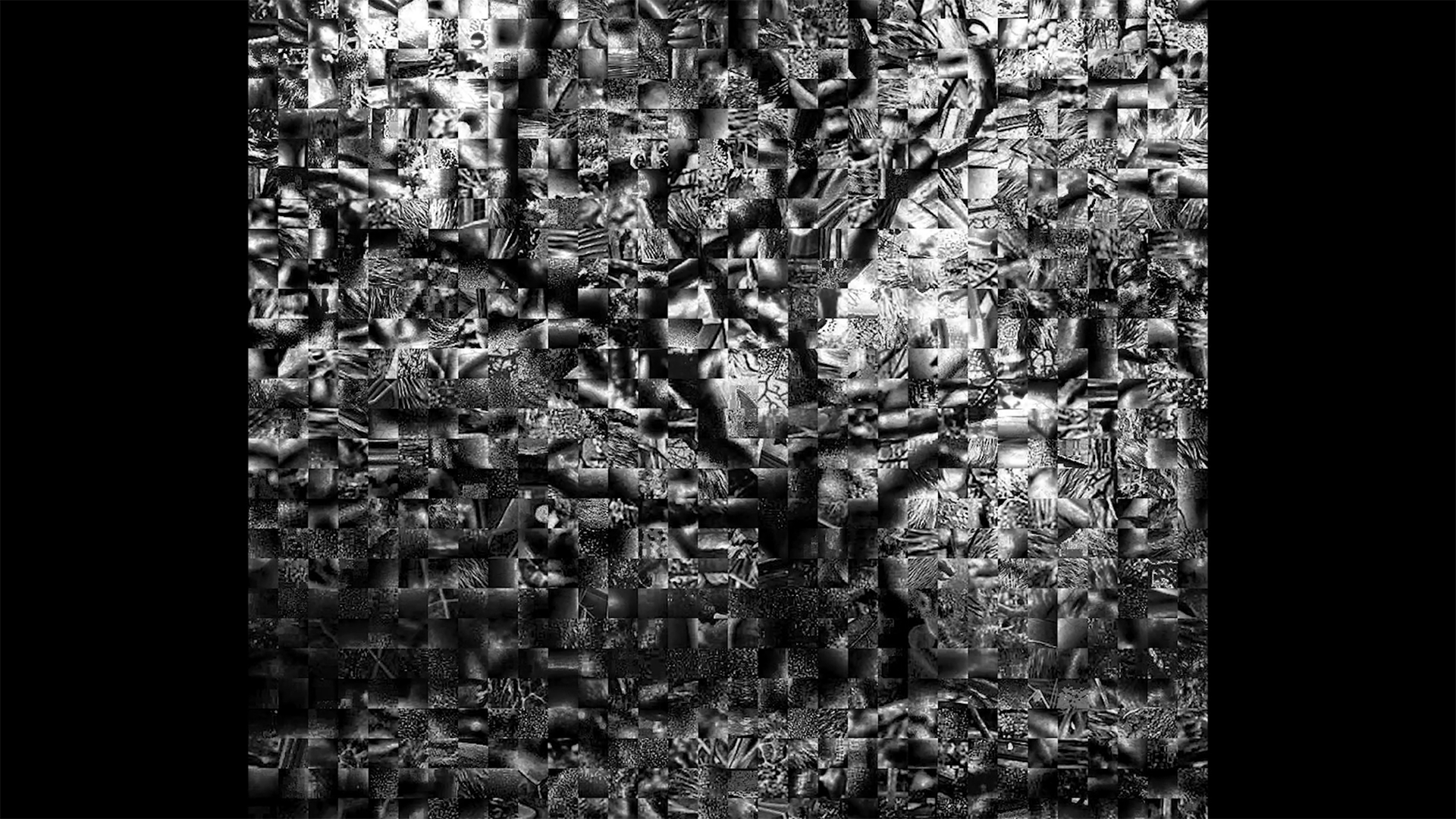}
  \end{minipage}
  \vspace{1mm}

  \begin{minipage}[b]{0.45\linewidth}
    \includegraphics[width=\linewidth]{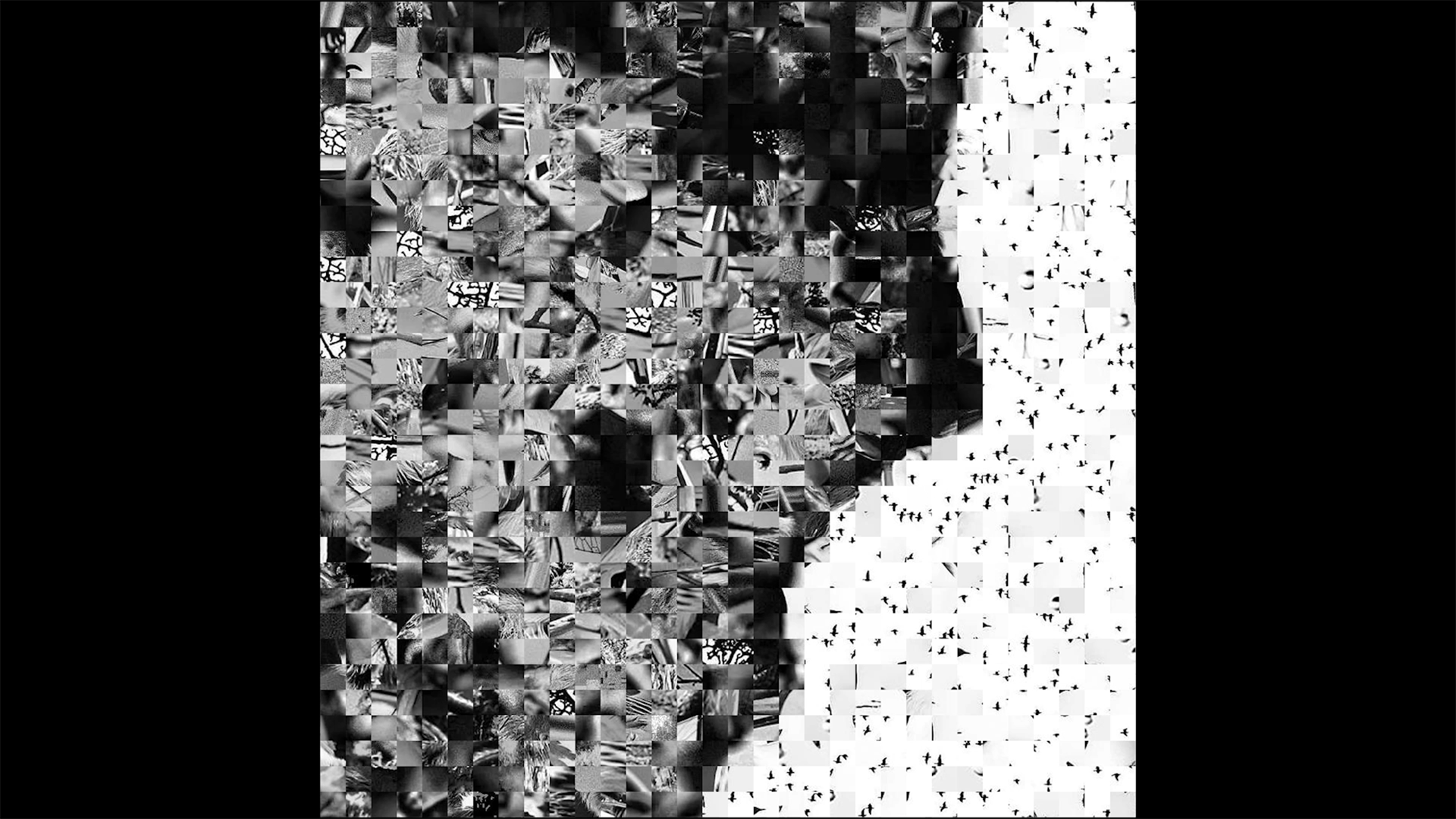}
  \end{minipage}
  \hfill
  \begin{minipage}[b]{0.45\linewidth}
    \includegraphics[width=\linewidth]{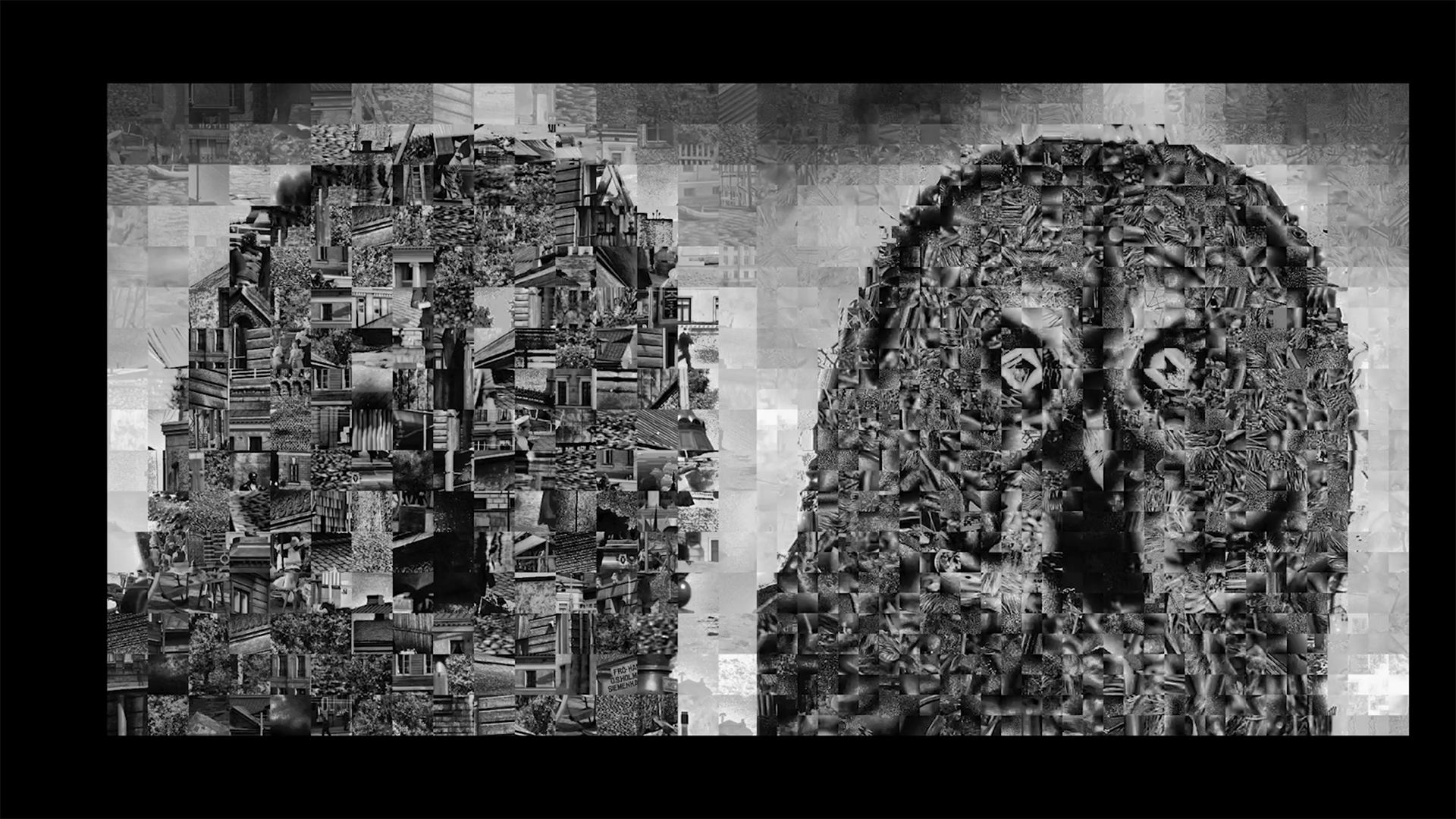}
  \end{minipage}
  \caption{Stills from Scene 8. First the natural image data set is introduced and then we see reconstructions of a few new images.}
\end{figure}

The final scene is a mixture of reconstructions of a few different images using the natural image data set. In addition, we play with some blending modes for different visual effects. 

\end{scenebox}

\vspace{0.5cm}

\subsection{Man on the moon}

\begin{figure}
    \centering
    \includegraphics[width=1\linewidth]{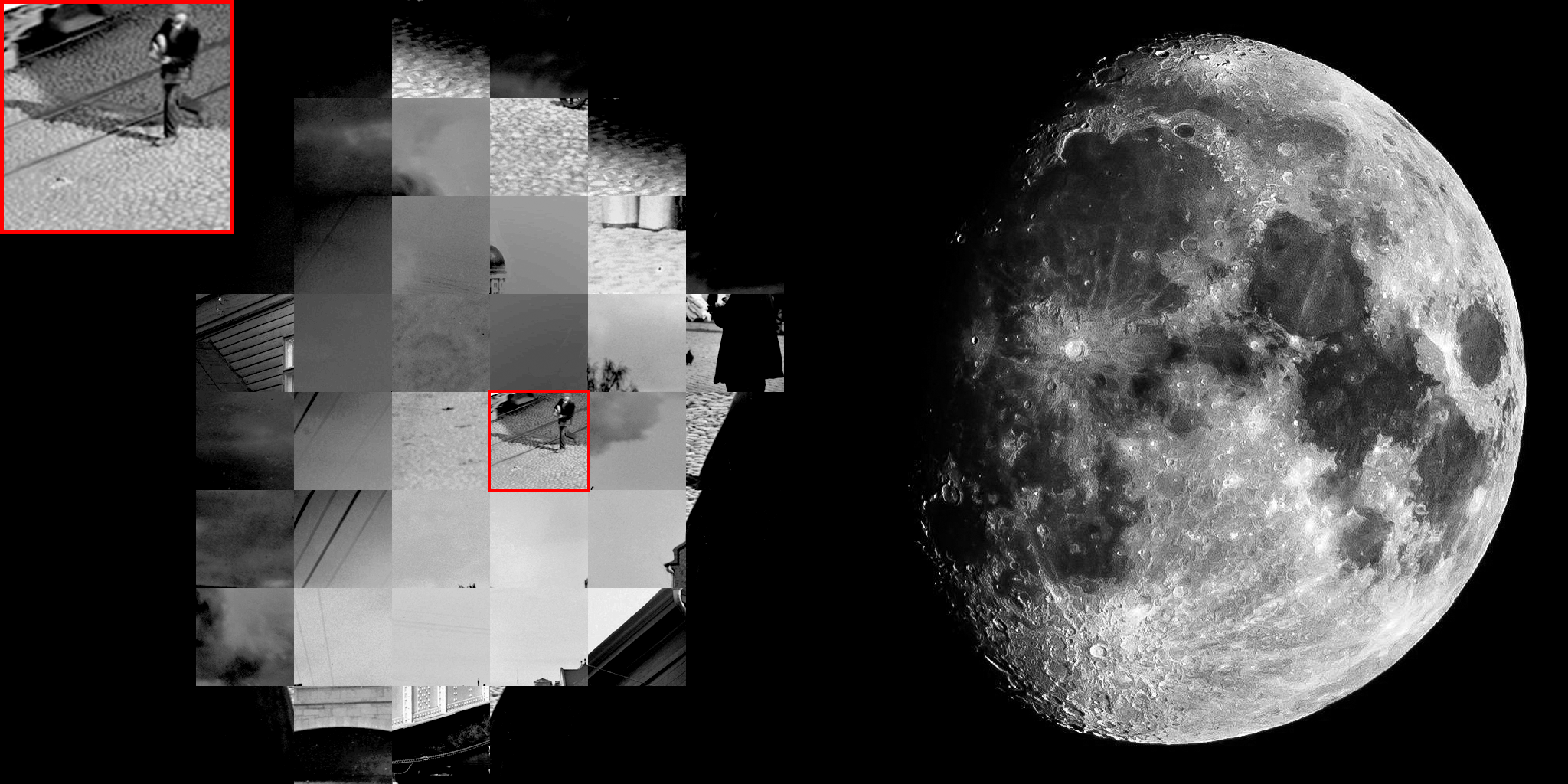}
    \caption{A reconstruction with $128\times128$ sized patches from the Old Helsinki data set. Left: the reconstruction with an additional upscaled view of the interesting patch (highlighted with a red square) with a man walking. Right: the original moon image of size $1024\times1024.$ }
    \label{fig:kuu_ukko}
\end{figure}

When reconstructing a moon image with $128\times128$ sized patches from the Old Helsinki data, we sometimes end up with interesting details like a man walking on the moon as seen in Figure \ref{fig:kuu_ukko}. This is one of the interesting aspects of this process. From one frame to another, we can have unexpected details pop up that stimulate reflection in the viewer. Depending on the speed of the changing frames in the final video, one can see glimpses of these possible cityscape details in a new and possibly unexpected setting.

\section{Discussion}

We described a way to recreate images using pieces of other images as a kind of mosaic with several choices (patch size, image data, cluster number) along the way. These choices affect not only the result, but also the time it takes to get to the result.
A large data set, large patches, and large number of clusters makes it heavy to compute. We reasoned that having a reasonably large data set and a large $k$ value would make the results better as long as the computation does not get too heavy and was manageable on the authors' laptops. There are several aspects of this process that could be modified that we want to mention here.

Instead of having equally sized patches for the entire image, we could start with larger patches and split the patches that don’t match well enough into 4 smaller patches similar to a quad tree \cite{Hunter1979quadtree}, until we have a match that is good enough for all patches according to a chosen metric.

For simplicity and a certain look we ended up using only square patches of the same size in one reconstruction, but one can extend this method to other patch shapes and adaptive patch sizes. We did experiment a bit with triangles and hexagons, but those results are out of the scope of this work, but perhaps worth refining. 

Possibly we could make some of the computations lighter by discarding 'boring' patches. We know and see that most of the patches in a set of natural images are uniform and don’t have a lot of edges and structure. We could just scrape a chunck of them and have fewer patches to deal with.

In the step of selecting random representatives from the clusters, we can end up selecting patches that are close to the boundary between two or possibly multiple clusters. This means that the difference between these two patches might be smaller than the difference between these patches and their respective clusters centroid. These borderline patches could be excluded from the final groups of patches that will be used in the reconstruction. In this way we would have better matching patches, but of course we might also lose some of the interesting varying details. Finding a good balance might be difficult.

As we are only using hundreds of images and hundreds of thousands of patches out of all possible patches, we can expect that the patch boundaries of the reconstructions don't always match so well and clearly visible boundary artifacts are introduced that look similar to severe JPEG image compression artifacts. In the results of this work we did not care about the boundary artifacts and embrace it as a feature of this method. In traditional mosaics, visible patch boundaries are also part of the style. Otherwise, we might not even suspect or know that the work has been composed of smaller pieces. Various methods of image stiching are available to mitigate these artifacts \cite{Wang2020} if chosen to do so.

We only use euclidian distance as the metric between patches as it is straightforward and works. Does it actually give the best distinction between patches? The use of a structural similarity metric was attempted to implement without a favorable outcome.

A color version of this method has been tested, but the clustering becomes heavier as the dimensionality increases. When moving from grayscale to color space, we also have to take into account that the color similarity is non-euclidian in RGB space. Moving to different color spaces such as Lab or YcbCr to do the computations is often done.

\section*{Acknowledgements}

The authors thank the Helsinki City Museum and the photographers whose images from their data base were used in this study.
Signe Brander,
Ragnar Nyberg,
Ivan Timiriasew,
Eric Sundström,
Olof Sundström,
Otto Johansson,
Eugen Hoffers,
Carl Otto Saxelin, and
A. E. Rosenbröijer.

\section*{Funding}

This work was supported by the Research Council of Finland (Flagship of Advanced Mathematics for Sensing, Imaging and Modelling grant 359182).

\bibliographystyle{tfs}
\bibliography{recomp_realities}

\appendix
\section{PCA and DCT comparison}

This appendix provides an additional comparison of the cluster centroids to components drawn from Principal Component Analysis (PCA) and the 2D Discrete Cosine Transform (DCT).

\begin{figure}
    \centering
    \includegraphics[width=1\linewidth]{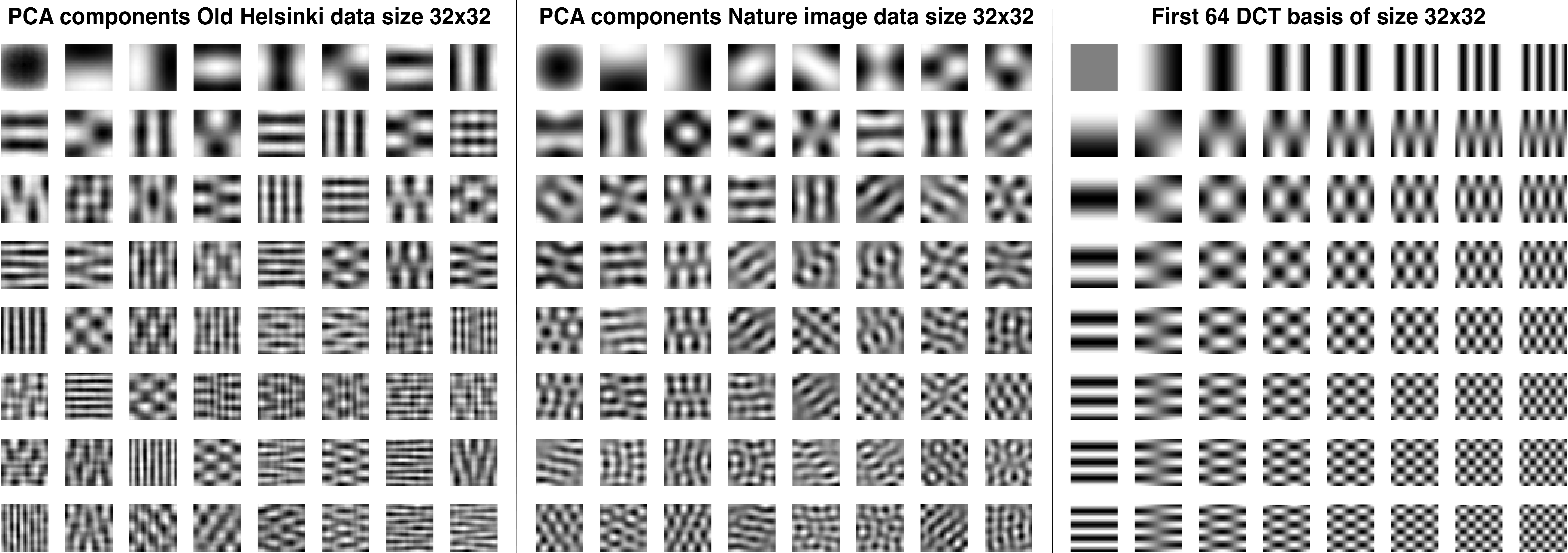}
    \caption{Left: The 64 largest PCA components for 32x32 sized patches from the Old Helsinki image set. Middle: The 64 largest PCA components for 32x32 sized patches from the nature image set. Right: The first 64 DCT components of size 32x32.}
    \label{fig:PCA_DCT_32size}
\end{figure}

Figure \ref{fig:PCA_DCT_32size} shows the PCA components of the $32\times32$ patches from both data sets alongside the first 64 of the 32x32 DCT bases for comparison. The PCA components are sorted from the most significant (top-left) to the least significant (bottom-right) based on their contribution to the dataset. Notice how the Old Helsinki dataset emphasizes more horizontal and vertical structures, while the Nature Image dataset highlights more diagonal and circular patterns, as we saw in the cluster centroids of same size shown in Figure \ref{fig:Cluster_means_k64_instaBW}. In contrast to the cluster centroids we find higher frequency components using PCA. This pattern of low-frequency dominance and gradually increasing frequencies is also observed in the DCT bases. This characteristic is expected, as DCT components are built from oscillatory functions that increase in frequency. This systematic arrangement and representation of image details underscore why DCT is the foundational method behind JPEG compression, widely adopted due to its effectiveness in representing images compactly.

 A key distinction in our approach compared to PCA or DCT is that we reconstruct images using real patches rather than combinations or weighted averages of multiple patches. Although this method may not always yield the most precise reconstruction, it fulfills our goal of authentically using genuine image patches rather than computationally derived averages.

\begin{figure}
    \centering
    \includegraphics[width=1\linewidth]{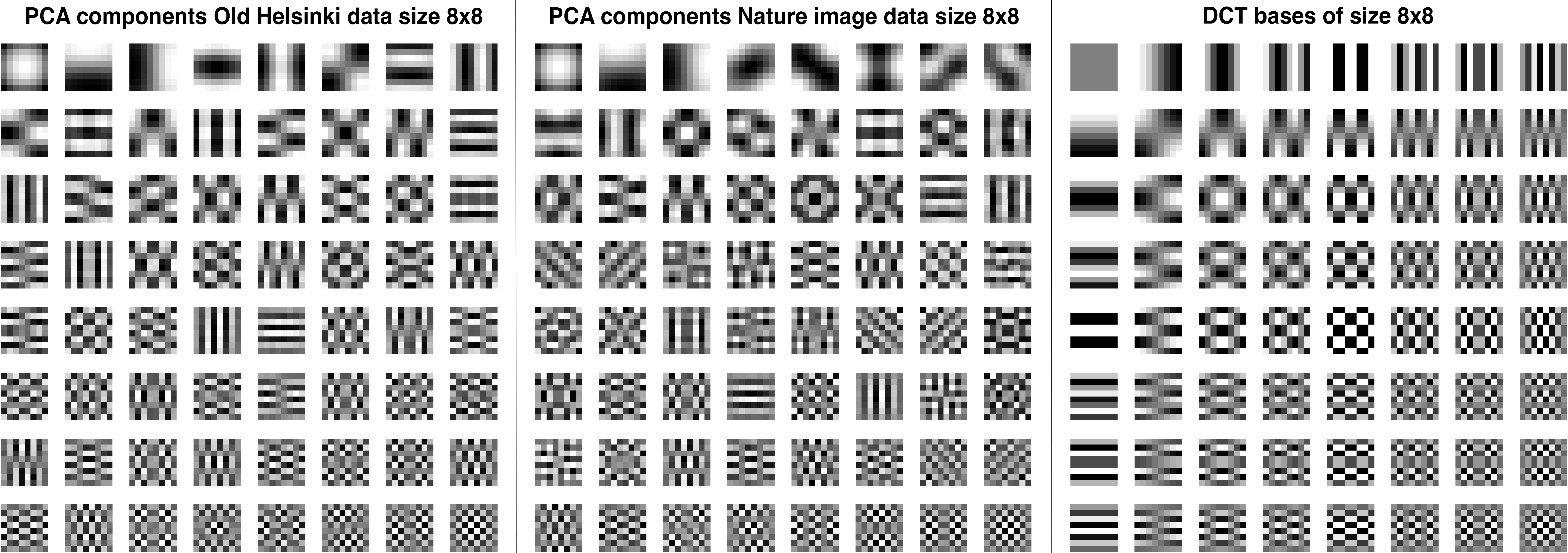}
    \caption{Left: The PCA components for 8x8 sized patches from the Old Helsinki image set. Middle: The PCA components for 8x8 sized patches from the nature image set. Right: 8x8 DCT bases.  }
    \label{fig:PCA_DCT_instaBW_8}
\end{figure}

\begin{figure}
    \centering
    \includegraphics[width=1\linewidth]{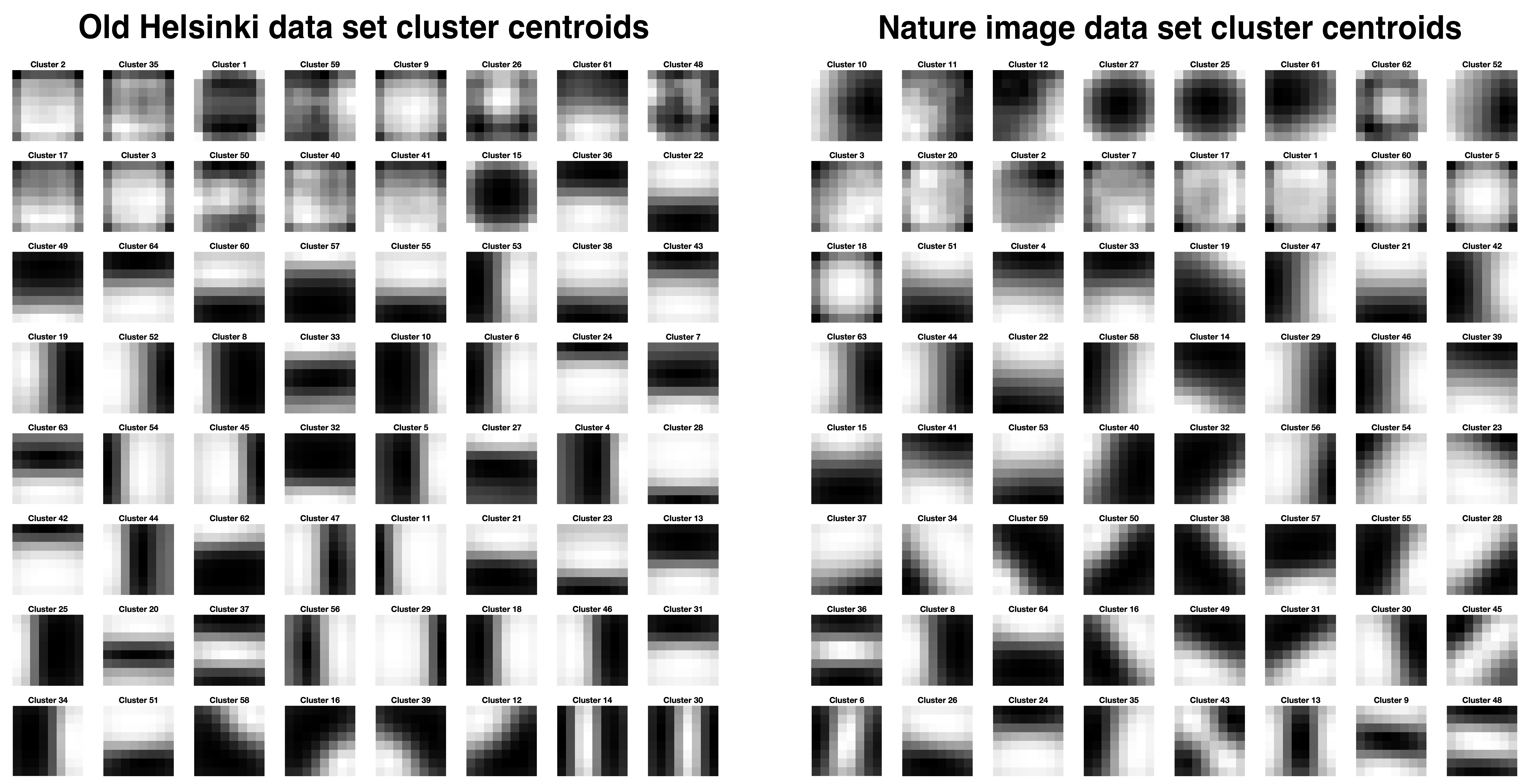}
    \caption{Cluster centroids from both data sets ordered from largest to smallest according to the amount of patches in the cluster (going row by row starting from the top-left). The patch sizes are $8\times8$ pixels and the amount of clusters is here 64 for both data sets. We see that the city data has much more horizontal and vertical edges than the nature set and vice versa the nature set seems to have more diagonal edges.}
    \label{fig:Cluster_means_k64_instaBW8x8}
\end{figure}

 Figures \ref{fig:PCA_DCT_instaBW_8} and \ref{fig:Cluster_means_k64_instaBW8x8} show the same breakdown for $8\times8$ patches. Again, if we look at the first PCA components of both data sets in \ref{fig:PCA_DCT_instaBW_8}, we see that the 3 first look identical. After this we have the same noticable difference in horizontal and vertical components versus more diagonal and circular ones. The cluster centroids in \ref{fig:Cluster_means_k64_instaBW8x8} reveal similar things, but the fact that the large amount of patches was grouped into only 64 clusters makes all clusters quite smooth and blurry compared to the larger ones.


 \end{document}